\def\year{2022}\relax
\documentclass[letterpaper]{article} % DO NOT CHANGE THIS
\usepackage{aaai22}  % DO NOT CHANGE THIS
\usepackage{times}  % DO NOT CHANGE THIS
\usepackage{helvet}  % DO NOT CHANGE THIS
\usepackage{courier}  % DO NOT CHANGE THIS
\usepackage[hyphens]{url}  % DO NOT CHANGE THIS
\usepackage{graphicx} % DO NOT CHANGE THIS
\usepackage{natbib}  % DO NOT CHANGE THIS AND DO NOT ADD ANY OPTIONS TO IT
\usepackage{caption} % DO NOT CHANGE THIS AND DO NOT ADD ANY OPTIONS TO IT
\usepackage{rotating}
\usepackage{xcolor}
\usepackage{booktabs}
\DeclareCaptionStyle{ruled}{labelfont=normalfont,labelsep=colon,strut=off} % DO NOT CHANGE THIS
\usepackage{algorithm}
\usepackage{bibentry}
\usepackage{algorithmic}

\usepackage{newfloat}
\usepackage{listings}
\usepackage{makecell}
\usepackage{float}
\floatstyle{ruled}
\newfloat{listing}{tb}{lst}{}
\floatname{listing}{Listing}
\title{Connecting Algorithmic Research and Usage Contexts: A Perspective of Contextualized Evaluation for Explainable AI}
\author{
    Q. Vera Liao \textsuperscript{\rm 1}\footnote{Part of the work was completed while the first and second authors were working at IBM Research},
     Yunfeng Zhang \textsuperscript{\rm 2},
     Ronny Luss\textsuperscript{\rm 3},
     Finale Doshi-Velez\textsuperscript{\rm 4},
     Amit Dhurandhar\textsuperscript{\rm 3}
}
\affiliations{
    \textsuperscript{\rm 1}Microsoft Research,
    \textsuperscript{\rm 2}Twitter Inc.,
    \textsuperscript{\rm 3}IBM Research,
    \textsuperscript{\rm 4}Harvard University
}
\begin{document}

\maketitle

\begin{abstract}
Recent years have seen a surge of interest in the field of explainable AI (XAI), with a plethora of algorithms proposed in the literature. However, a lack of consensus on how to evaluate XAI hinders the advancement of the field. We highlight that XAI is not a monolithic set of technologies---researchers and practitioners have begun to leverage XAI algorithms to build \textit{XAI systems} that serve different usage contexts, such as model debugging and decision-support. Algorithmic research of XAI, however, often does not account for these diverse downstream usage contexts, resulting in limited effectiveness or even unintended consequences for actual users, as well as difficulties for practitioners to make technical choices. We argue that one way to close the gap is to develop evaluation methods that account for different user requirements in these usage contexts. Towards this goal, we introduce a perspective of contextualized XAI evaluation by considering the relative importance of XAI evaluation criteria for prototypical usage contexts of XAI. To explore the context dependency of XAI evaluation criteria, we conduct two survey studies, one with XAI topical experts and another with crowd workers. Our results urge for responsible AI research with usage-informed evaluation practices, and provide a nuanced understanding of user requirements for XAI in different usage contexts. 
\end{abstract}

\noindent \section{Introduction}
The wide adoption of AI technologies in high-stakes domains, coupled with the proliferation of inscrutable ``black-box'' AI models, has spurred great interest in explainable AI (XAI) in academia and industry. Each year, hundreds of papers proposing various XAI algorithms are published. Unfortunately, a lack of consensus on what constitutes good explanations hinders the advancement of the field and real-world adoption of XAI. While practitioners recognize the value of explainability, they grapple with tremendous challenges in making appropriate choices of XAI techniques and creating effective XAI systems~\cite{bhatt2020explainable,liao2020questioning,hong2020human}.  Researchers, especially those in the human-computer interaction (HCI) community, have begun to explore diverse XAI systems (e.g.~\cite{kaur2020interpreting,xie2020chexplain}). These studies pointed out that explainability is not a monolithic concept, and what users need to be explained varies across different types of systems and user tasks such as debugging a model, judging the reliability of model outputs, assessing regulatory compliance, or learning about a domain. 

%From supporting model development, allowing auditing for responsible AI, to engendering appropriate trust in end users, the field of XAI has come to bear many high hopes.

%The question of what makes an explanation ``good'' has been repeatedly asked in philosophy, social science and cognitive science. It is difficult to answer due to a wide variability of structures, stances, and patterns in people's everyday explanations. 

This recognition of the context-dependency of explanation ``goodness'' resonates with social science literature studying human explanations~\cite{mueller2019explanation,vasilyeva2015goals}. An explanation is often conceptualized as an attempt for the explainer to fill the gaps in the understanding of the explainee. Therefore its goodness should be relative to this understanding gap, which is determined by both the explainee's current understanding and the necessary understanding to achieve their given objective. Social science literature distinguishes different objectives for people to seek explanations, including predicting future events, diagnosis, assigning blame, resolving cognitive dissonance, rationalizing actions, and aesthetic pleasure~\cite{lombrozo2012explanation,keil2006explanation,lombrozo2006structure}.

However, this context-dependent nature of explainability is not well-acknowledged in current XAI research. There is a fundamental disconnect between algorithmic research and downstream usage contexts. Algorithmic research is often not motivated by well-defined needs of intended users~\cite{miller2017explainable}. In fact, the intended use is often not made explicit, despite growing efforts in encouraging AI researchers to articulate the downstream impact of their research.  This disconnect has been recognized to cause pitfalls of XAI methods~\cite{liao2021human,ehsan2021explainability}---unintended harmful consequences for users such as lacking actionability, cognitive burden and over-reliance on AI. 

This disconnect is reflected in the dominant practices of how XAI algorithms are evaluated, which can profoundly shape the field. A major camp of XAI evaluation focuses solely on algorithmic criteria such as faithfulness and stability~\cite{alvarez2018towards,carvalho2019machine}, which are inadequate to capture the satisfaction of ``users in context''~\cite{hoffman2018metrics}. To move towards a rigorous science based on empirical evidence, ~\citet{doshi2017towards}'s foundational work called for application-grounded evaluation---with real humans and by the success of target tasks. When the resource is limited, compromises can be made with simplified or proxy tasks. However, how to design simplified or proxy evaluation tasks that capture the essential requirements for different application contexts remains an open question. Despite increasing effort to conduct human-subject studies to evaluate XAI algorithms, popular evaluation methods are often devoid of usage contexts and under-specify the criteria. Some asked participants to judge which explanation is ``better''~\cite{jeyakumar2020can}, without specifying an end-goal to ground the definition of ``better''. Another approach is to use a simplified task of ``simulatability test'' by asking participants to simulate the model output based on the explanation~\cite{hase2020evaluating,lucic2020does}. A recent study by~\citet{buccinca2020proxy}  pointed out that such tasks have limited evaluative power to predict the success in real tasks, as they do not capture what users need to be explained or how they process the explanations.

Our work is motivated to close these gaps. We propose to contextualize the evaluation of XAI by considering the relative importance of evaluation criteria for prototypical usage contexts of XAI systems. We use the term ``contexts'' to refer to situations where explanations are sought for distinct user objectives, which can appear in different applications. In Section 3, we first contribute taxonomies of XAI evaluation criteria and prototypical usage contexts by synthesizing related literature. In Section 4, we empirically explore the context-specific ranking of these evaluation criteria by surveying XAI experts (from both HCI and AI communities) and crowd workers as target users of an AI application. As a first step towards contextualized evaluation of XAI, we use a retrospective approach by surveying people's opinions to systematically explore this large problem space. We believe our results can contribute to four grounds:
\vspace{-0.3em}
\begin{itemize}
    \item Our work aligns with responsible AI research---by explicating the context-dependency of evaluation criteria, we encourage XAI researchers to articulate the appropriate usage contexts for the algorithms they develop. 
    
    \item By capturing the desired properties of XAI, our results also inform how to develop XAI techniques and systems that serve different usage contexts.

    \item Methodologically, we contribute a scenario-based survey approach to elicit crowd workers' opinions on the relative importance of evaluation criteria.
    
    \item By examining usage-informed evaluation criteria and the gaps in existing XAI evaluation practices, including nuanced differences between how end-users and XAI experts perceive the criteria differently, we call out implications to reflect on the common evaluation criteria as value-laden choices that will shape the field.
    
\end{itemize}{}

\section{Related work}
%\vera{Do we miss any important areas of related work?}

\subsection{XAI methods and usage contexts}

%While the origin of ``XAI'' can be traced back to expert systems in the 1980s~\cite{clancey1983epistemology}, the field has seen a surge in the past five years, fueled by public concerns over the increasing use of incomprehensible ``opaque-box'' AI. 

Although there is a lack of consensus on the definition of explainability, XAI works share a common goal of making AI models understandable by people. Many recent papers surveyed the \textit{how} (methods) and \textit{why} (objectives) of the XAI field~\cite{guidotti2018survey,arrieta2020explainable,adadi2018peeking,gunning2019xai,lipton2018mythos,gilpin2018explaining,molnar2020interpretable}. At a high level, the technical landscape of XAI can be divided into two camps: 1) directly interpretable models; 2) adopting “black-box” models such as deep neural networks and large tree ensembles, and then using post-hoc techniques to generate explanations.  ~\citet{guidotti2018survey} differentiate between model explanations (i.e. global explanations), local outcome explanations and model inspection. Under these categories, there are different explanation methods, such as leveraging features or examples to explain; and for each method, many techniques have been developed with differences in computational properties.

While some criticized that XAI techniques tend to be ``developed in a vacuum''~\cite{miller2019explanation}, the field largely recognizes that there are ``no one-fits-all solutions''~\cite{arya2019one} for the rich application opportunities of XAI. Recent works began to characterize the main user groups of XAI and postulate their different requirements~\cite{preece2018stakeholders,hind2019explaining,arrieta2020explainable}, including model developers, regulatory bodies, business owners, direct end-users, and impacted groups. Other studies identified similar roles based on empirical studies of ML practitioners~\cite{bhatt2020explainable,hong2020human}. 

%Some also defined user groups by their knowledge or expertise, such as AI novices, AI experts, and data experts~\cite{mohseni2018multidisciplinary}. 

A recent work by ~\citet{suresh2021beyond} points out that these frameworks lack granularity to distinguish between attributes of the users and their objectives to seek explanations. For example, people in any role may want to assess model biases or improve the model at certain usage points. Thereby~\citet{suresh2021beyond} define stakeholders' knowledge and their objectives as two components that cut across to characterize the space of user needs for explainability. The authors further propose a multi-level typology to characterize XAI users' objectives, ranging from long-term goals (building trust and understanding the model), immediate objectives (debug, ensure regulatory compliance, take follow-up actions, justify actions influenced by a model, understand data usage, learn about a domain, contest model decisions), and specific tasks to perform with explanations.

%(assess prediction reliability, detect mistakes, understand information used by the model, understand feature influence, understand model strengths and limitations). 

 Our definition of \textit{XAI context} is similar to the ``immediate objective'' in Suresh et al.'s framework as characterizing a situation for which a user seeks explanations\footnote{Our taxonomy is largely consistent with Suresh et al.'s with some variations:  we consider ''adapting control'' as requiring ``understanding data usage''; we focus on immediate interactions and leave out the downstream objectives such as ``contesting model''.  }. In Section 3, we synthesize a list of prototypical XAI usage contexts with additional prior works reviewed. We choose to focus on XAI contexts following the objective-dependent stance of explainability in the social science literature~\cite{lombrozo2012explanation,keil2006explanation}, along with a practical goal of informing context-specific design of XAI. We acknowledge that other factors such as user characteristics can further vary the evaluation criteria. Our work moves beyond the existing effort of characterizing the problem space to informing concrete context-specific requirements, aiming to provide actionable guidance for the evaluation and design of XAI. 

\vspace{-0.5em}
\subsection{XAI evaluation
}

Our approach is informed by a bulk of prior research on XAI evaluation. Our focus is on ``XAI evaluation criteria''---normative properties of ``what constitute good explanations''~\cite{hoffman2018metrics}---which should be distinguished from outcome measurements of using XAI such as identification of bugs or improvement of decisions. For normative criteria, earlier works focused on \textit{model intrinsic} criteria such as faithfulness (how well the explanations approximate the original “black-box” model’s decisions) and stability (how consistent the explanations are for similar cases). Recent work began to define criteria that capture aspects of human perception of ``good'' explanations, such as comprehensibility, actionability, and interactivity~\cite{carvalho2019machine,sokol2020explainability}. 

However, we must recognize that there are varying priorities, even trade-offs, between these criteria depending on the context. For example, while an ML engineer might demand faithful explanations to engage in model debugging tasks, a layperson using a decision-support AI may be willing to sacrifice some degree of faithfulness for compactness. We can find support for this context-dependency in recent HCI works studying different XAI systems. For example, model debugging tools often integrate detailed local and global explanations~\cite{narkar2021model,hohman2019gamut}. Decision-makers were found to have less desire for global explanations during time-constrained decisions, and prefer less distracting information~\cite{xie2020chexplain}. For AI capability assessment, a study hinted on that example-based explanations may have an advantage to expose users to the AI limitations~\cite{buccinca2020proxy}.

As mentioned, a widely cited XAI evaluation framework is the taxonomy by~\citet{doshi2017towards}, proposing three categories with decreasing specificity and cost: application-grounded evaluation with humans and real tasks, human-grounded evaluation with humans and simplified tasks, and functionally grounded evaluation with no humans and proxy tasks (e.g., quantifying with some formal definition of human desired property). Our perspective builds on this framework and extends it by calling out the need to design simplified and proxy evaluation tasks based on evaluation criteria that are important to the target application context. For example, in a decision-making context with AI assistance, a common user objective is to have appropriate reliance, knowing when not to rely on the AI when it is likely to err. Recent work highlights the communication of model \textit{uncertainty} as a desired property of XAI~\cite{wang2021explanations,carvalho2019machine}. However, this criterion cannot be captured by the commonly used simplified evaluation task, ``simulatability test''~\cite{buccinca2020proxy}. We speculate that methods that directly measure the success of uncertainty communication---such as by whether people can correctly judge if a model prediction is uncertain based on the explanation, or by quantifying the correlation between some notion of uncertainty salience in the explanations and ground-truth uncertainty---would make more effective evaluation tasks for this usage context. 

 Lastly, researchers have begun to explore a growing number of XAI systems~\cite{hohman2019gamut,zhang2020effect,kaur2020interpreting,xie2020chexplain}. However, the design choices and evaluation measurements used were largely inconsistent and ad-hoc, which hinders the development of scientific knowledge about human-XAI interaction and principled design guidelines. By exploring the prioritized evaluation criteria, i.e. desired properties, of explanations for prototypical XAI usage contexts, we also aim to inform the design choices and evaluation practices for researchers and practitioners working on XAI systems.

\section{Taxonomies Development}
We conducted a literature search to consolidate taxonomies of evaluation criteria and prototypical usage contexts of XAI.  With Google Scholar (retrieved by December 2020), we used search terms ``explainable/ interpretable AI/ML'' + ``evaluation/ assessment/ metrics'' for the former, and + ``goal/ objective/ context/ motivation/ use case'' for the latter. After an initial review, we focused on a subset of papers with taxonomies or frameworks proposed. As enumerated below, we note that both taxonomies are necessarily incomplete given the fast advancement of the field. However, we consider them as sufficiently comprehensive, and contend our methodology can be used to extend the results.

\subsection{XAI Evaluation Criteria}
As mentioned, we focus on normative explanation ``goodness" criteria, which can be further differentiated between \textit{model intrinsic} properties of explanations (e.g., faithfulness), and \textit{human-centered} properties that reflect the perception of the explainee (e,g., comprehensibility). Model-intrinsic properties can usually be measured by computational metrics while human-centered properties are best measured by human responses with questionnaires or behavioral measures. This differentiation is not binary, as it is possible to devise proxy measures to assess human-centered criteria. For example, while the ``compactness'' criterion (being succinct and not overwhelming) is contingent on the explainee's perception, it is possible to define some notion of ``information units'' to quantify the compactness of XAI output~\cite{abdul2020cogam}. Lastly, we differentiate between an evaluation \textit{construct} (\textit{what} criterion) and an evaluation \textit{method} or metric (\textit{how} to measure). While our studies focus on the constructs, as we enumerate the criteria below, we also discuss existing methods to measure them, if any, or potential directions to develop new methods.

%Hoffman et al.~\cite{hoffman2018metrics} and Mueller et al.~\cite{mueller2019explanation} consider goodness as ``properties of the explanation that are often
%taken as axiomatic assumptions about what constitute good explanation'', while common outcome measurements of human-XAI interaction include mental model improvement, task performance, user satisfaction, and trust in AI. Our focus is the former, as our aim is to directly inform desired properties for ``good'' explanations.

%The above conceptualization is relevant to Doshi-Velez and Kim's three-level taxonomy of XAI evaluation approaches~\cite{doshi2017towards}: while outcome measurements are ideally used with application-grounded evaluation, goodness criteria can be used in human-grounded evaluation (simplified tasks measured by human responses) and functionally -grounded evaluation (proxy tasks using computational metrics). That being said, it would also be desirable to measure goodness criteria in application-grounded evaluation~\cite{hoffman2018metrics} to gain a better understanding of \textit{what aspects} of the XAI features need improvement~\cite{hoffman2018metrics}. The context-dependent choice of goodness criteria, which our work aims to inform, can guide such choices, as well as design of simplified and proxy tasks for target applications. 

The papers we reviewed are distributed in the AI and HCI communities~\cite{sokol2020explainability,carvalho2019machine,yeh2019fidelity,alvarez2018towards,murdoch2019interpretable,schneider2019personalized,mohseni2018multidisciplinary,guidotti2018survey,miller2017explainable,jesus2021can,lakkaraju2020fool,hancox2020robustness,gilpin2018explaining,doshi2017towards,kulesza2013too,kulesza2015principles,hoffman2018metrics,hsieh2020evaluations}.  We found most criteria are covered by~\citet{carvalho2019machine} and an ``explainability requirements fact sheet'' by~\citet{sokol2020explainability} (criteria under ``usability requirements'' instead of developer requirements). We develop our list based mainly on these papers, supplemented with additional items and definitions from others (sources are cited for each criterion below)  We arrived at the following list of evaluation criteria, with definitions used in the survey in \textit{italic}. 

%We encourage readers to review the ``training material'' in Appendix C, with examples of good and bad explanations according to each criterion, which was used to introduce end-user survey participants to these definitions.

%Most existing works focus on metrics that quantify intrinsic properties of output generated by a given XAI algorithm, such as fidelity and stability~\cite{}. Some also propose recipient-dependent measures such as comprehensibility, interactivity, and coherence with the recipient's prior knowledge~\cite{}. Recent HCI works also discuss the importance for explanations to communicate model (un)certainty and be actionable for down-stream tasks~\cite{}.

%\textcolor{purple}{FD: connect to the goal, what intermediate metrics may be key for a downstream task}

%\textcolor{red}{RL: We should discuss that there's overlap between various criteria.}
\vspace{-0.3em}
\begin{itemize}
    \item \textbf{Faithfulness}: \textit{The explanation is truthful to how the AI gives recommendations}~\cite{alvarez2018towards}, also referred to as fidelity~\cite{carvalho2019machine,ras2018explanation} or soundness~\cite{kulesza2013too}. It is a commonly used criterion, especially to evaluate post-hoc explanations, with computational metrics proposed in the literature~\cite{alvarez2018towards,yeh2019fidelity}.
    
    \item \textbf{Completeness}: \textit{The explanation covers all components that the AI uses to give recommendations, or can generalize to understand many AI recommendations}~\cite{sokol2020explainability,kulesza2013too,gilpin2018explaining}, also referred to as representativeness~\cite{carvalho2019machine}. It is considered an orthogonal aspect to faithfulness for an explanation to accurately reflect the underlying model. According to~\citet{sokol2020explainability}, it can be quantified by metrics that reflect the coverage or generalizability across sub-groups of a data set.

    \item \textbf{Stability}: \textit{The explanation remains consistent for similar cases I ask about}~\cite{carvalho2019machine,alvarez2018towards}, also referred to as robustness~\cite{hancox2020robustness}. While in some cases it can be at odds with faithfulness (if the model decision itself is unstable), stability is argued to be important if the goal is to understand not just the model, but true patterns in the world~\cite{hancox2020robustness,alvarez2018towards}. Prior works proposed several metrics ~\cite{alvarez2018towards,hsieh2020evaluations}.

    \item \textbf{Compactness}: \textit{The explanation gives only necessary information and does not overwhelm}, also referred to as parsimony~\cite{sokol2020explainability,ras2018explanation}. It reflects the design principles of ``providing appropriate details''~\cite{mueller2019explanation,kulesza2015principles}.  This criterion can be at odds with completeness. Computational metrics with some notion of information units or cognitive chunks have been proposed~\cite{abdul2020cogam,doshi2017towards} to proximate compactness, but capturing the essence of ``necessary information'' may require task-specific definitions or subjective responses.

    \item \textbf{(Un)Certainty (communication)}: \textit{The explanation reflects the (un)certainty or confidence of the AI in its recommendations}~\cite{carvalho2019machine}. Recent studies underscored this criterion to support appropriate reliance on AI~\cite{zhang2020effect,bansal2021does}. To our knowledge, there is no established method to measure this criterion. It is possible to devise survey scales to measure certainty perception or computational metrics with some definition of certainty representation, and then compare them against the true certainty or correctness of model predictions.
    
    \item \textbf{Interactivity}: \textit{The explanation is interactive and can answer my follow-up questions}~\cite{sokol2020explainability}. Recent work considers interactivity as a necessary requirement for XAI, given the diverse knowledge gaps people have~\cite{liao2020questioning,weld2019challenge}. While being interactive can encompass many capabilities, we adopt a broad definition of allowing users to specify their explainbility needs by asking questions interactively. Behavioral measurements and computational metrics of interactivity can be conceived based on the scope of interactions the XAI can support.
    
    \item \textbf{Translucence}: \textit{The explanation is transparent about its limitations, for example, the conditions for it to hold}~\cite{carvalho2019machine}. It is referred to as contextfulness in~\citet{sokol2020explainability}, which discussed multiple types of explanation limitations such as ambiguity and a lack of generalizability. While under-explored in the current literature, it is a criterion that could critically impact people's trust in the explanation itself. Translucence should be measured with regard to exposing the true limitations of the explanations.
    
    \item \textbf{Comprehensibility}: \textit{The explanation is easy to understand (e.g., intuitive, taking less time to understand)}~\cite{carvalho2019machine}, also referred to as clarity~\cite{ras2018explanation}, understandability~\cite{guidotti2018survey}, and explicitness~\cite{alvarez2018towards}. It is explainee-dependent and best measured by subjective or behavioral measures such as understanding correctness and speed, although it is possible to proximate with some quantifiable latent properties~\cite{doshi2017towards}. While comprehensibility and compactness can correlate, the latter is more about cognitive workload and does not guarantee easiness to understand.

    \item \textbf{Actionability}: \textit{The explanation helps me determine follow-up actions to achieve my goal for the task}~\cite{sokol2020explainability}. This criterion is contingent on the explainee's goal, thus should be measured by goal-specific subjective responses or behavioral measurement on the success of user goals specific to the evaluation task.

    \item \textbf{Coherence}: \textit{The explanation is consistent with what I already know about the domain}~\cite{sokol2020explainability}. \citet{miller2019explanation} posits that a desirable property of explanation is coherence with the explainee's prior knowledge. This is a subjective criterion and should ideally be measured by subjective responses.
    
    \item \textbf{Novelty}: \textit{The explanation provides new or surprising information that I otherwise would not know}~\cite{sokol2020explainability,carvalho2019machine}. In some contexts, such as scientific discovery, the utility of explanation may stem from providing novel information to guide users. Novelty is also highly subjective and should be measured by self-reported responses.

    \item \textbf{Personalization}: \textit{The explanation is tailored to my needs and preferences, e.g. level of details, language style, etc.}~\cite{sokol2020explainability}. Lastly, we include a criterion that focuses on the \textit{communication style} according to one's preferences. Satisfaction with personalization should ideally be measured by subjective responses.

\end{itemize}{}
\vspace{-0.3em}
These criteria are \textit{not} all orthogonal. Some can be correlated (e.g., compactness and comprehensibility) or involve trade-offs (e.g., coherence and novelty, coherence and faithfulness, completeness and compactness, stability and faithfulness). Such relationships indeed underline our motivation to identify their priorities in different usage contexts.
\begin{figure*}
  \centering
  \includegraphics[width=1.8\columnwidth]{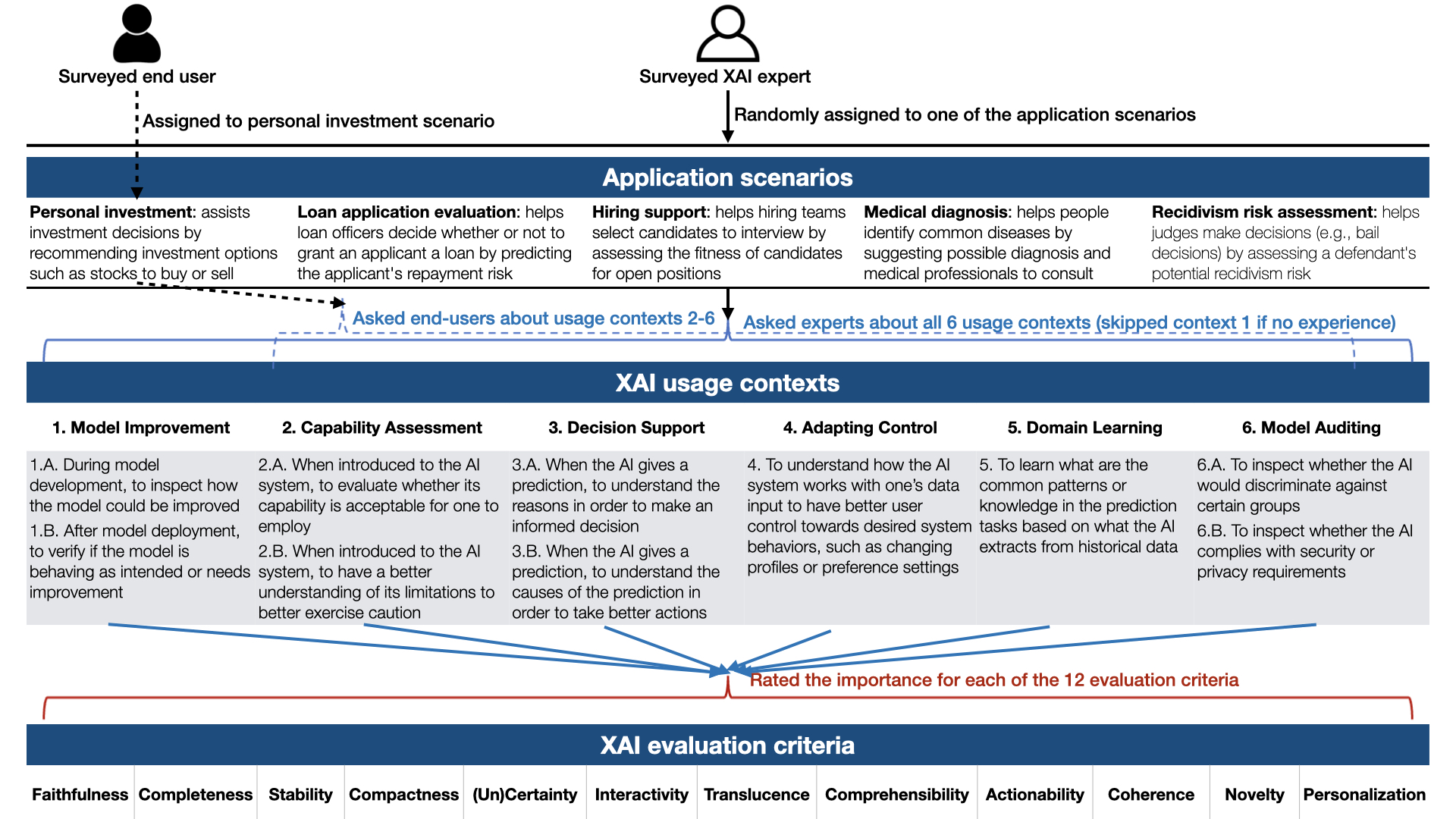}
   \vspace{-1em}
  \caption{Experiment procedure and descriptions of AI applications and usage contexts. Each participant experienced one application, asked about all (5 for end-users) contexts, for each of which they rated the importance of all evaluation criteria  }~\label{fig:experiment}
   \vspace{-2.5em}
\end{figure*}

\subsection{XAI Contexts}

Our literature search on usage contexts of XAI discovered two bodies of work. First, many survey or perspective papers discussed conceptually the objectives for people to seek explanations~\cite{doshi2017towards,adadi2018peeking,samek2019explainable,guidotti2018survey,mohseni2018multidisciplinary,arrieta2020explainable,molnar2020interpretable,lipton2018mythos}. For example, beside~\citet{suresh2021beyond} reviewed earlier,  \citet{chen2021towards} emphasizes the need to make choices of XAI techniques according to use case goals and proposes a taxonomy of XAI use cases including model debugging, promoting trust in a model, assisting scientific discovery, assisting human decision-making, and providing actionable recourse. Another body of work summarizes common user objectives with XAI by empirically studying real-world use cases~\cite{hong2020human,bhatt2020explainable}. We choose to mainly build on~\citet{liao2020questioning}, which summarizes user objectives with XAI by studying 16 real-world AI systems, supplemented with additional definitions from others. Figure~\ref{fig:experiment} shows our list of usage contexts: Model Improvement, Capability Assessment, Decision Support, Adapting Control, Domain Learning, and Model Auditing. We intentionally kept these definitions general to avoid priming participants' judgment by a specific form of explanation. Following Liao et al., under some categories we consider sub-categories that share a similar goal but have different focuses (e.g., assess AI capability versus limitations). These sub-categories were randomly selected to present to participants.

\vspace{-0.5em}
\section{Survey studies}

We solicited people's opinions on the relative importance of XAI evaluation criteria with two scenario-based survey studies. First, we surveyed XAI experts from both the AI and HCI communities for multiple types of AI applications. We then surveyed a selected group of crowd workers as target users of an XAI application assisting personal investment. We first describe the design and analysis of the two surveys separately, and then discuss results from both together.

\vspace{-0.5em}
\subsection{Study 1: XAI Expert Survey}
\paragraph{Survey Design}
The survey presents different contexts in which a stakeholder seeks explanations from an AI application. To explore whether results vary by domain, we created the same set of contexts for five decision-support AI applications. We started with an AI application scenario database created by~\citet{lubars2019ask}, then narrowed down our choices based on the following criteria: 1) It should be feasible for current AI so there is a shared expectation of how the AI works; 2) It should be commonly known so participants can easily imagine the scenario; 3) The decision is high-stakes so the need to understand AI likely presents.  We selected the 5 AI application scenarios described in Figure~\ref{fig:experiment}: personal investment, loan application evaluation, hiring support, medical diagnosis, and recidivism risk assessment.

%\begin{itemize}
    
%\item \textbf{Personal investment}: Assists investment decisions by recommending investment options such as stocks to buy or sell%
%\item \textbf{Loan application evaluation}:  helps loan officers decide  whether or not to grant an applicant a loan by predicting the applicant's repayment risk
%\item \textbf{Hiring decisions}: Helps hiring managers and teams select candidates to interview for open positions by assessing the fitness of each candidate 
%\item \textbf{Medical diagnosis}: Helps people identify common diseases by %suggesting possible diagnosis and the right medical professionals to consult
%\item \textbf{Recidivism risk}: Helps judges make decisions (e.g., bail decisions) by assessing a defendant's potential recidivism risk
%\end{itemize}

The experiment procedure is illustrated in Figure~\ref{fig:experiment}. Each expert participant was randomly assigned to questions regarding one of the 5 AI applications. The application was introduced in the instructions and presented at the top throughout the survey, including its functions, the data to train it, and example features of the model (See the survey page screenshot in Figure~\ref{fig:survey} in Appendix A). Descriptions of the 5 applications are also in Appendix A. On each survey page, participants were presented with one of the 6 XAI usage contexts in Figure~\ref{fig:experiment} as a task that a stakeholder performs (see the Task part of Figure~\ref{fig:survey} in Appendix A). We customized the task descriptions for the given application. Participants finished survey pages for all 6 XAI contexts in Figure~\ref{fig:experiment} (one of the sub-categories, if any, was randomly chosen), with the exception of Context 1: Model improvement. Participants were asked whether they were experienced in model development, and were only presented with Context 1 if their answer was yes. All but two expert participants answered yes. For each task, participants were asked to rate the importance of 12 evaluation criteria that explanations should satisfy for them to accomplish the task. All responses are five-point Likert-type from ``not important'' to ''very important''.

%The description includes the AI application's goal, function, the type of data to train it, and examples of features used.  

%To entice participants to also provide rankings instead of tied ratings, after they submitted the ratings for each AI task, we asked them to rank the top five criteria they rated (or more than five if there was a tie). The top rated criteria were provided in the order of their ratings, randomly ordered if there was a tie, and participants could drag-and-drop to change the order, as shown in Figure~\ref{fig:rank}. They were also asked to rate their confidence in understanding the task and providing appropriate ratings, and optionally provide an open-text comment. 

 In summary, our survey introduces three independent variables for the analysis: each participant was assigned to questions regarding one of the five \textit{AI applications}, performed judgment tasks with all \textit{XAI contexts}, and rated all \textit{evaluation criteria} for each context. The importance ratings for the evaluation criteria are used as the dependant variable.

Participants were given a consent form and instructions in the beginning, and asked to rate their expertise in XAI and confidence in addressing the AI application at the end. For each context, we asked an optional question to comment on the reasons of their ratings. 5 participants left at least one comment. We also asked participants to leave their contact information if they are willing to answer follow-up questions. We emailed those who did to solicit comments on their ratings and 5 responded. In the result section, we discuss some observations from these qualitative data.

%Participants produced two sets of results that will be analyzed as dependent variables separately: the importance ratings and assigned ranks for the evaluation criteria. We gathered both to prevent the potential noises of tied ratings.

%\vera{TBD: should we skip mentioning the ranking part altogether? we can also mention including rank analysis in the appendix, but then we need to double the pairwise comparion tables}

%\vera{TODO @Amit, would you be able to draw a sketch or find a similar example? This is common in HCI study description so unclear what is needed to communicate it} \textcolor{blue}{AD: Should add a nice figure here I think that in one look showcases the design i.e. 6 contexts, 12 metrics, rate (Likert) and rank, 2 groups (XAI experts and end users), etc.}

\vspace{-0.5em}

\paragraph{Participants: XAI experts}
We adopted a snowball sampling strategy by initially contacting 25 researchers in different organizations, who have published substantially on XAI. We strove to have a balance of AI and HCI researchers in our initial contacts to ensure both algorithmic and user perspectives were presented. We asked them to forward the recruiting message to other people working on the topic, emphasizing that this survey targets XAI experts. Participants' information based on the exit survey is summarized in Appendix C. Notably, 22 out of the 35 participants consider themselves to be ``very knowledgeable'' on the topic of XAI and none has ``no knowledge'' (another 11 knowledgeable, and 2 somewhat knowledgeable). 32 have the job title of researcher (the remaining 3 are data scientists).

%In total, we gathered completed survey responses from 35 people. Among them, 22 considered themselves to be ``very knowledgeable'' about XAI, 10 as ``knowledgeable'' and 3 as ``somewhat knowledgable''. 5 indicated to be ``very confident'' in addressing the given AI application, 12 as ``confident'', 13 as ``somewhat confident'' and 5 as ``slightly confident''. 32 identified their job titles as ``researchers or scientists'', while the remaining 3 as ``data scientists'' or ``ML engineers''. 10 self-identified as female, 23 as male, and 2 declined to answer. 

%8 of them are under age 30, 15 between 30-39, 1 between 40-49, 6 between 50-59, 2 above 60, the remaining 3 declined to answer. 

\vspace{-0.5em}

\paragraph{Expert Survey Analysis} We performed a mixed-effects regression analysis on participants' ratings, by including the AI application (between-subjects), XAI context (within-subjects) and evaluation criterion (within-subjects) as fixed-effects variables, and participant as a random-effects variable\footnote{We acknowledge the potential loss of information by modeling Likert-type responses using parametric tests~\cite{owuor2001implications}. We also modeled the data with an ordinal logistic regression, and the patterns largely hold. However, the results are more difficult to interpret.  We use parametric tests for presentation clarity.}. We found the main effect of XAI context ($F(5,2106)=3.351, p=0.004$), evaluation criterion ($F(11,2106)=2.589, p=0.003$), and the interactive effect between the two ($F(55,2106)=3.351, p=0.038$) to be significant. The main effect of AI application is not significant ($F(4,30)=1.108, p=0.371$), nor any of its interactive effects. These patterns indicate that experts' relative ratings of evaluation criteria were significantly varied by the XAI contexts, but not the types of AI application.
 
 %We found similar patterns with a regression analysis on the ranks, with significant main effect of XAI context ($F(5,2106)=2.250, p=0.050$), criterion ($F(11,2106)=2.132, p=0.016$), and their interactive effect ($F(55,2106)=1.381, p=0.034$), as well as a non-significant effect of AI application ($F(4,30)=1.848, p=0.146$). These patterns indicate that the relative ranks and ratings of evaluation criteria were varied by the XAI contexts, but not the types of AI application.

%\textcolor{purple}{\textbf{DISCUSS} FD: adjust type of language based on the type of p-value correction, okay for things to be suggestive}

%\textcolor{red}{RL: We should summarize at the beginning that AI Scenario was not significant. }

Given the \textit{non-significance of AI application}, we merged data across the five applications. For each XAI context, we conducted an omnibus-ANOVA analysis on the ratings, with evaluation criterion as a within-subject variable. The main effect of evaluation criteria is significant ($p<0.001$) for all six XAI contexts. We then conducted pairwise post-hoc t-tests for all contexts, using false discovery rate (FDR) adjustment to adjust p-values for multiple comparisons. The first part of the table in Figure~\ref{tab:results} presents the mean ratings provided by expert participants. The last column presents the overall mean ratings by combining ratings for all 6 XAI contexts.  We include all the post-hoc analyses statistics in Appendix D. We acknowledge that given the relatively small sample size, the individual post-hoc analysis may not have enough power to be conclusively insignificant. In Section 4.3 we discuss patterns, and highlight comparisons that are statistically significant even with the small sample.

%Table~\ref{tab:experts-rank} presents similar results with the median ranks in increasing order.

%We conducted similar, but non-parametric tests for ranks: we started with a Friedman test on ranks and found the main effect of evaluation criterion to be significant ($p<0.001$). We then conducted pairwise post-hoc Conover's tests using FDR adjustment. All statistical results are in the Appendix.

%The comparative positions of ranks and ratings are largely consistent, with all but two criteria (the positions of fidelity and (un)certainty for Adapting Control are reversed for ratings) varied no more than two ranks.

\subsection{Study 2: End-User Crowd Survey}

\subsubsection{Survey design}
%\textcolor{red}{RL: Re "general population": maybe say target users are easier to target in MTurk than for the other scenarios.}
We chose to use only one scenario to conduct the end-user survey---personal investment. Compared to the other applications, its target user group are more general and thus easier to recruit crowd workers as suitable target users. Furthermore, the analysis of the XAI expert survey showed no significant effect of the application scenario. The ``Model Improvement'' context was removed since most workers do not have relevant experience. 

The survey design was largely similar to the expert survey, except that we also designed a \textit{training task} to introduce crowd workers who are not familiar with XAI to the definitions of the 12 evaluation criteria. As presented in Appendix B, using a movie recommender scenario, each training task page introduces the definition of one evaluation criterion, shows two explanations, and asks participants to choose which better exhibits the given criterion. After submitting the response, the page indicates the right choice and highlights the corresponding part of the explanation that exhibits the criterion considered. In the survey, participants could hover over a criterion and a pop-up window would show the training task to help them recall the definition. Participants were told that they would only be qualified for the survey if they pass at least half of the 12 training tasks. If they fail, they would be compensated with a base payment of \$2 USD. If they finish the survey they would receive an additional \$4 USD in compensation. Most participants took 20-40 minutes to complete the survey.

\vspace{-0.5em}
\paragraph{Participants: End-users}
We added a filter ``Financial Asset Owned'' on Mechanical Turk to target workers with experience in financial investment.
In total, we gathered completed survey responses from 37 people, and excluded data points from 5 workers who took less than 10 minutes (less time than any experts). The background information of the 32 participants is shown in Appendix C. In contrast to participants of the expert survey, none consider themselves to be ``very knowledgeable'' on XAI, and 27 to be ``no knowledge'' or ``somewhat knowledgeable''. The two groups are comparable in other dimensions such as confidence in addressing the AI application and demographic attributes. Possibly due to the filter used, these participants appear to be relatively tech-savvy compared to the general population, which we consider as consistent with a user population that would use an AI assisted investment application.
%Among the remaining 32 participants, 5 considered themselves to be ``knowledgeable'' with the topic of XAI,  16 as ``somewhat knowledgeable'', and 11 as ''no knowledge''. 4 indicated to be ``very confident'' in addressing the given AI application, 15 as ''confident'', 12 as ``somewhat confident'' and 1 as ``slightly confident''. 8 identified their job titles as ``manager'', 4 as ``software engineer'' 3 as ``data scientist'', 2 as ``designer'', 1 as ''researcher'', and the remaining 14 as ``other''. 14 self-identified as female, 18 as male. 

%4 of them are under age 30, 16 between 30-39, 7 between 40-49, 2 between 50-59, 3 above 60.

\vspace{-0.5em}
\paragraph{End-User Survey Analysis} We started by examining whether the experts' and end-users' ratings are significantly different. We performed a mixed-effect regression on ratings by including XAI context, evaluation criterion and participants group (experts/end users) as fixed-effects variables, and participant as random effects. We found the main effect of group ($F(1,65)=0.436, p=0.511$) as well as the interactive effect of XAI context, evaluation criterion, and group ($F(44,3835)=0.970, p=0.529$) to be non-significant; but the two-way interaction between evaluation criterion and group is significant ($F(11,3835)=2.689, p=0.002$). These patterns suggest that there is no significant difference between how XAI experts and end-users considered the relevant importance of XAI evaluation criteria in different XAI contexts, but there might be differences in their overall ratings of evaluation criteria, which we discuss later.

%Similarly, with the same analysis on ranks, we found the main effect of group ($F(1,65)=0.190, p=0.664$) and the interactive effect of XAI context, criterion and group ($F(44,3835)=0.990, p=0.490$) to be non-significant, but the two-way interaction between criterion and group is significant ($F(11,3835)=2.807, p=0.001$). These patterns suggest that there is not a significant difference between XAI experts and end users in considering the relevant importance of XAI evaluation criteria in different XAI contexts, but there might be difference in their overall ranking of evaluation criteria, which we will discuss in detail in Section X by comparing the two groups' results.

An ANOVA analysis was conducted on the ratings for each XAI context from end-users. We found the main effect of evaluation criterion to be significant for every XAI context ($p<0.001$), then conducted post-hoc pairwise comparisons. Detailed statistics are in Appendix D. The second part of the table in Figure~\ref{tab:results} presents the mean ratings of evaluation criteria for each context provided by end-user participants. The last column presents the overall mean ratings by combining all five XAI contexts (no ``Model Improvement'' in the end-user survey). We also merged data from experts and end-user surveys and conducted the same ANOVA and pair-wise analyses, with the combined mean ratings presented in the third part of Figure~\ref{tab:results} and the detailed statistics are in Appendix D.
%Table~\ref{tab:users-rank} presents similar results with the median ranks.

\vspace{-0.5em}
\subsection{Discussions of survey results}
\begin{figure*}
  \centering
  \includegraphics[width=2\columnwidth]{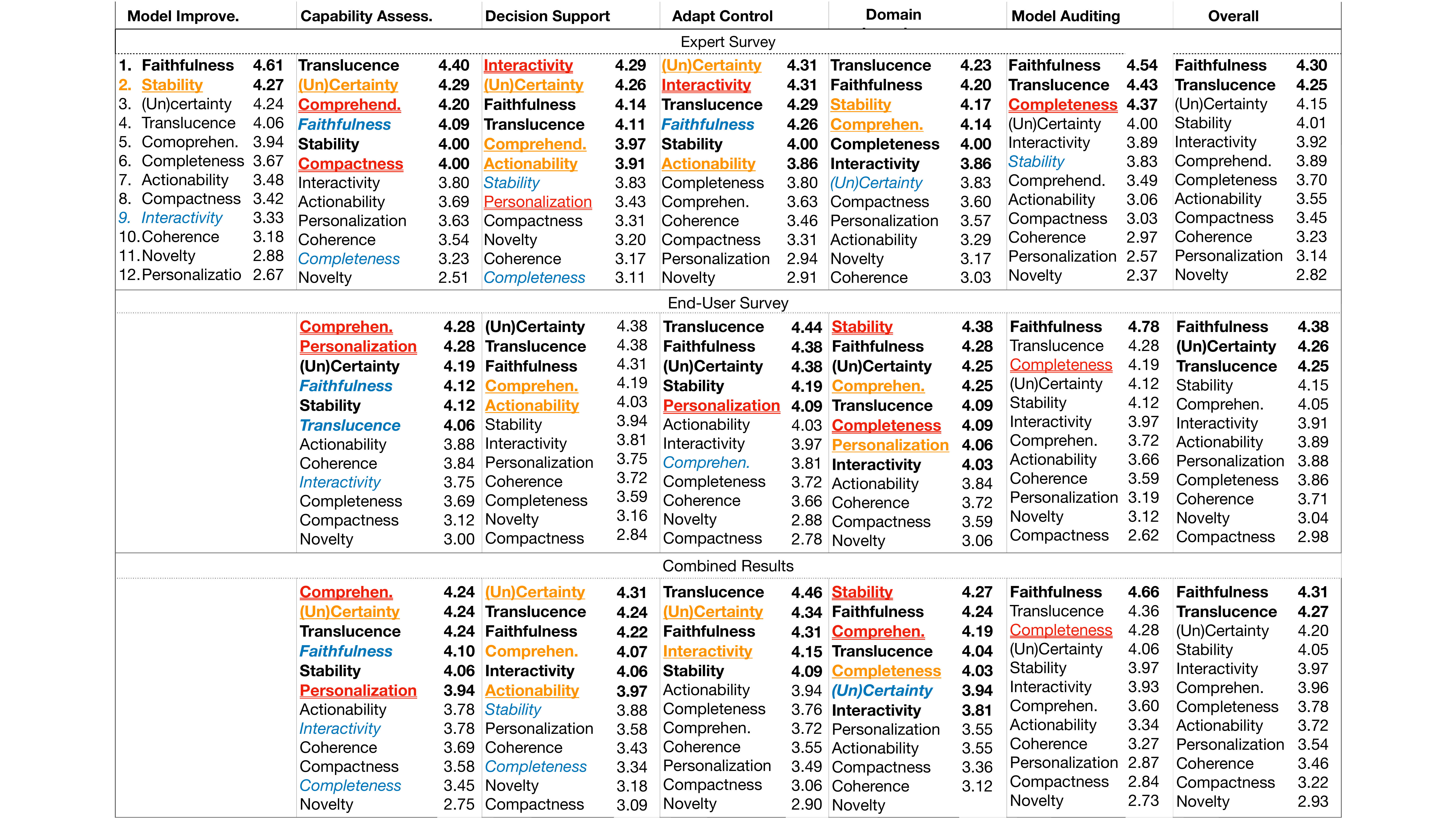}
  % \vspace{-1em}
  \caption{Average importance ratings of evaluation criteria by XAI experts, end-users and combined.  Bold fonts are ``top groups''. Double-underlined red ones increase more than 2 ranks compared to the overall ranking; Single-underlined orange ones increase 1-2 ranks and also enter the top group; Italic blue ones decrease more than 2 ranks. }~\label{tab:results}
  % \vspace{-2.5em}
\end{figure*}
We refer to the table in Figure~\ref{tab:results} to discuss the results from both surveys. Each column presents participants' mean ratings for evaluation criteria for one usage context, in decreasing order. The last column presents the overall mean ratings by combining data across all contexts. The first part of the table shows ratings from the expert survey; the second part is from the end-user survey; and the last part combines data from the two surveys. As mentioned, the three-way interaction between usage context, evaluation criterion and participant group is non-significant, meaning the experts and end-users did not have a statistically significant difference in how they consider the relative importance of evaluation criteria for different contexts. Therefore, below we start with patterns in the combined results, and then discuss the nuanced differences between experts' and end-users' ratings. 

Fonts and color coding are used to visually highlight a few patterns in the table in Figure~\ref{tab:results}. First, we bold the ``top group'' for each context. While our sample size might have lacked enough power to conclude all insignificant pairwise comparisons, we adopt the following heuristic: for each XAI context, we look at the pairwise comparisons with the top criterion, and consider criteria that had less than marginal significance ($p>0.100$) to be closer to the top criterion than others, as in the top group. Second, we color code the relative importance of evaluation criteria that \textit{vary in different XAI contexts}. For each set of results (experts/end-users/combined), we use the criteria positions in the last column of overall ratings as a reference and highlight prominent differences in other columns. We visually highlight the differences with the following heuristics: 1) we code criteria that increase more than 2 ranks in \textit{{\color{red}red}} and \underline{\underline{double-underlines}}; 2) we code criteria that increase 1-2 in ranks and also enter the ``top group'' in \textit{{\color{orange}orange}} and \underline{underline}; 3) we code criteria that decrease more than 2 ranks in \textit{{\color{blue}blue}} and \textit{italic} . In short, the red and orange colors indicate a relative increase in position for a usage context compared to others, while the blue color indicates a relative decrease. Based on the table in Figure~\ref{tab:results}, in the following, we first enumerate the patterns shown for each usage context, then summarize some take-aways.
\vspace{-0.5em}
\subsubsection{Patterns by usage contexts}

\textit{Across all contexts},  faithfulness was considered the most important criterion, followed by translucence, uncertainty and stability. Both experts and end-users also considered interactivity and comprehensibility to be relatively important. Novelty, compactness and coherence were rated consistently at the bottom. While experts also rated personalization to be of little importance, end-users considered it slightly more important. Below we elaborate on the results for each usage context.

\textit{Model improvement.} Only experts provided ratings for this context. Faithfulness and stability were rated as the top criteria. Stability was especially deemed more important than for other contexts, while interactivity was considered less of a requirement than for other contexts. These patterns suggest that to support model improvement, such as debugging, it is important to \textit{provide faithful and stable explanations for users to inspect the true causes of model issues}. Since users are often ML experts to perform this task, they may be more receptive to complex or information-rich explanations~\cite{narkar2021model,hohman2019gamut} without the need for interactive inquiries, as evidenced in P20's response in explaining their ratings:``\textit{It is vital to know how much can be relied upon the explanation. Hence, the need for fidelity, stability and translucence... I deem interactivity as relatively less important due to the assumption that an expert attempts to improve it and thus has the skills and competencies to get to the bottom of some model behavior.} ''

\textit{Capability assessment.} For the combined results, comprehensibility, uncertainty, translucence, faithfulness, stability and personalization were rated as top criteria.  Comprehensibility, uncertainty and personalization were rated more important for capability assessment than for other contexts, while faithfulness was comparatively lower. Interactivity and completeness were considered less important. End-users especially rated comprehensibility and personalization high, while experts rated compactness higher than for other contexts. These patterns suggest that to support capability assessment, such as during users' onboarding stage, it is important to provide \textit{easy-to-understand explanations to help users make an efficient assessment}, even at some cost of faithfulness; \textit{being transparent about model uncertainty and explanation limitations} are also important; but \textit{complete and interactive explanations may not be necessary as users prioritize intuitiveness and efficiency}. A nuanced difference is that end-users considered personalization, while experts considered compactness, as important to make an efficient assessment. P33 explained their willingness to compromise faithfulness and completeness for comprehensibility: ``\textit{Fidelity is still important, just didn't seem to be as important... because some omission on details may be beneficial for first-comers. Same goes for completeness, as users may just want to assess the model within the context of their own use.}''

\textit{Decision support.} Uncertainty, translucence, faithfulness, comprehensibility, interactivity and actionability  were rated closely high in importance. Uncertainty, comprehensibility and actionability were deemed more important than for other contexts, while stability and completeness were less important. Interestingly, experts and end-users disagreed on the importance of interactivity, with experts rating it as a top criterion. These patterns suggest that for decision support, it is important that explanations should \textit{communicate uncertainty of model predictions and the limitations of explanations}; it is also important to ensure the explanations are \textit{easy to understand and actionable for the decision goal}, which may also require \textit{being interactive} to help users reach the necessary understanding. Since decision support often focuses on explaining individual predictions, stability (whether it applies to similar cases) and completeness (whether it generalizes or covers all decision components) are less important. As P19 commented: ``\textit{Completeness is more important when I need an understanding of the whole system... here this is mostly not the case. Instead it is very important to know the certainty, so I can take the prediction with `a grain of salt' if necessary and be extra careful. Interactivity is very important when I have the feeling I need to `dig deeper' ... Comprehensibility is so that I learn to integrate the explanation into my work routine without losing efficiency}''.

\textit{Adapting control.}  Translucence, uncertainty, faithfulness, interactivity and stability were rated as top criteria. It suggests that to support a better understanding of how the system works with one's data to make adjustment, explanations should \textit{be transparent about the model uncertainty and explanation limitations}, and also \textit{be faithful and stable to allow users to discover and change the true causes of sub-optimal system behaviors}. Interestingly, comprehensibility was rated lower than for other contexts, suggesting a willingness to invest time for this task that may only be performed occasionally but has a lasting impact, as P10 commented: ``\textit{I think that comprehensibility can be lower here because the user has time to stop and think for a while}.''  Experts rated interactivity higher for this context, but not end-users; instead end-users rated personalization higher. They suggest the two groups may have conceptualized the requirements differently. Since adapting control requires users to locate precise causes they can control, such as changeable preference settings, experts may have envisioned it as an interactive process to narrow down the causes, while end-users may have preferred the system to proactively adapt to their desired level of details. P19 (expert)'s comment provides support: ``\textit{Having low certainty tells me where I need to tweak it further. Translucence is important because I do not want to make any changes based on information that was not even relevant. Stability is important because if I instruct the system to behave differently I expect it to generalize this behavior to other similar cases. Interactivity empowers me to dig deeper in cases where I am puzzled why a certain behavior occurred and may test some what-if hypothesis. }''

\textit{Domain learning.} Stability, faithfulness, comprehensibility, translucence, completeness, uncertainty and interactivity were rated as top criteria. Stability, comprehensibility and completeness were rated higher than for other contexts, while uncertainty was rated less important, mainly due to experts' opinions. These patterns suggest that to support learning, explanations should \textit{be stable, faithful and relatively complete to help people discover true patterns of the domain}; it should also be \textit{easy-to-understand}, as users may not be ML experts or interested in the model details. P20 commented on the importance of stability, faithfulness, comprehensibility and translucence: ``\textit{These four properties are in my opinion needed to enable a user to ascertain actionable knowledge about a domain. This matches with the literature on education and training---humans require reliable explanations they can understand to acquire new knowledge.}''

\textit{Model auditing.} Only faithfulness is in the top group, as a clear winner for this usage context, followed by translucence. Meanwhile, completeness was rated higher for this context than for others. These patterns suggest that in order to support model auditing, explanations should be first and foremost \textit{faithful to the underlying model}, \textit{transparent about the explanation limitations}, and \textit{cover the decision components completely, to allow accurate and comprehensive auditing}, as supported by P2's reasoning: ``\textit{faithfulness and translucence are important because I need to ensure that I am diligent when checking for legal-compliance. Hence I need explanations that are truthful to the model (rather than basing my assessment on potentially wrong information) and cover all components of the system being assessed.}''

\vspace{-0.5em}
\subsubsection{Implications for designing and evaluating XAI} We reflect on the results above and discuss some implications. 

\textit{Faithfulness is critical, but can be compromised in some contexts.} Both experts and end-users recognize that faithfulness is the most important criterion across the board, especially for model improvement and auditing to enable diagnosing the true model issues, validating the field's current focus on quantifying faithfulness~\cite{alvarez2018towards,yeh2019fidelity}. However, participants considered it somewhat compromisable in other contexts such as capability assessment. Importantly, these observations imply that whether post-hoc explanations or explanations with less faithful details are desirable depends on the usage contexts.

\textit{Translucence and uncertainty are important requirements but not well-supported in current XAI research.} A striking result is that participants consistently rated translucence and uncertainty communication high for all contexts, despite currently limited XAI approaches that support these criteria (though there has been a long line of research on directly quantifying uncertainty~\cite{bhatt2021uncertainty,ghosh2021uncertainty}). We highlight such user needs for the field to focus on, including defining evaluation tasks that capture whether an XAI technique can support users to make correct judgments of model uncertainty and explanation limitations.

\textit{Requirement for comprehensibility is prominent when efficiency and reducing cognitive load matter.} Comprehensibility was generally rated important but less so for adapting control, model auditing and model improvement, all of which require more engagement with inspecting and changing the model. In participants' qualitative responses, its high priority was frequently mentioned together with the importance of efficiency, or users' unwillingness to spare time or cognitive resources, such as when gauging system capability or prediction reliability in everyday use.

\textit{Stability and completeness are important when having a generalizable understanding matters.} Stability is a criterion that reflects consistency and robustness of explanations for similar cases. Completeness reflects how well the explanation covers all components (e.g., features, decision paths) so users can apply the knowledge to understand not one but many predictions. We found the relative ranks of the two often have correlated changes: both were rated more important for domain learning, but less important for decision support. We postulate that both criteria are desired when users aim to identify generalizable and robust patterns. Curiously, completeness was rated low for capability assessment. We suspect the reason is that completeness is seen as requiring more time and cognitive effort, as evidenced by the often negative correlation in the rank movement between completeness and comprehensibility.

\textit{Requirements for interactivity and personalization are perceived differently by XAI experts and end-users.}  Experts rated interactivity high for decision support and adapting control, but end-users did not consider it as important; while experts consistently rated personalization at the bottom, end-users considered it a top criteria for capability assessment, adapting control and domain learning. While caution should be exercised interpreting the results as the two groups may have different conceptualizations, including what is technically feasible, of these criteria, these differences are worth reflecting on and being further explored. Interactivity is increasingly discussed in XAI literature as an important direction for the field~\cite{miller2019explanation,weld2019challenge}, which would allow user inquiries to guide the provision of desired explanations to achieve their goal. This trend may have influenced experts' positive ratings on interactivity. It does not necessarily mean well-designed interactivity is undesired by end-users, but at a conceptual level, they might have preferred the system to take the initiative to provide desired explanations directly, such as through personalization. The difference between the two groups' ratings on personalization may reflect a gap between what experts of the field prioritize and what end-users actually need.

\textit{Requirement for actionability is neutral.}  The reason could be that the success of supporting users' end goals is better captured by other more specific criteria (e.g., comprehensibility for capability assessment) rather than the generic statement about ``supporting follow-up actions''. Given that actionability is also hard to operationalize without running an application-grounded evaluation, we postulate that when designing simplified and proxy evaluation tasks, it may be more productive to focus on the other top criteria.

\textit{The importance of explanation novelty, coherence and compactness is unclear.} Despite being discussed in the literature as important usability requirements~\cite{sokol2020explainability}, these criteria were consistently rated at the bottom. It is possible that some other top criteria correlate with them, but better capture people's desired properties. For example, comprehensibility could be a more informative criterion than compactness for the field to focus on. As shown in the example used in our training task, an example-based explanation could be more intuitive, but not more compact compared to a feature-based explanation. As for novelty and coherence, it is possible that personalization could better capture the desired relation between explanations and ones' prior knowledge. We also believe more research is required to explore the roles of these criteria in actual user interactions beyond the retrospective approach we took.

\section{Discussions and Future Work}
Our central thesis is that to close the gaps between algorithmic research and their effectiveness in deployment necessitates explicitly considering different requirements in different downstream usage contexts. Towards this goal, we introduce a perspective of contextualized evaluation.  We empirically show that the relative importance of explanation ``goodness'' criteria varies across prototypical usage contexts of XAI. We suggest ways for future work to use our results.

\noindent\textbf{Articulating the appropriate usage contexts of XAI algorithms}. This is one area where contextualized evaluation is critically needed for responsible research and enabling others, including practitioners, to make appropriate use of research outputs. It is possible to use our results to perform a weighted analysis for that purpose.  Assuming measurement scores of a given XAI technique can be obtained for all the evaluation criteria, one can use the importance ratings in Figure~\ref{tab:results} to inform a set of weights for each usage context, then generate a weighted sum of evaluation score for each context respectively. Contexts with low weighted scores should be cautioned against applying the given technique. For example, an example-based explanation may score relatively low in stability and completeness, but high in uncertainty communication and comprehensibility. Its weighted scores would be high for decision-support but low for model improvement and auditing, which is consistent with observations in HCI studies~\cite{cai2019effects,buccinca2020proxy,dodge2019explaining}. Analyses can also be performed with a set of existing XAI techniques to benchmark desired scores for different usage contexts.

\noindent\textbf{Selecting and optimizing XAI for a given usage context}. Another common scenario where our results can be applied is to optimize XAI features for a given usage context. A researcher or practitioner's task may be to select from available XAI techniques, or to evaluate a current XAI technique and identify its shortcomings to make improvements. The results in Figure~\ref{tab:results} can help guide which evaluation criteria to focus on. Especially when performing a full application-grounded evaluation is beyond the resource allowed, one can focus on the top criteria identified for the given usage context using simplified or proxy metrics, or consider how to further optimize these top desired properties. For instance, when developing an XAI feature to support domain learning, one may want to prioritize using XAI techniques that score high in stability and faithfulness. One may also need to improve a current XAI feature by enhancing comprehensibility and translucence. 

\noindent\textbf{Reflecting on what evaluation criteria the field prioritizes as value-laden choices}. Our work has another important goal---empirically examining the desired criteria in usage contexts and the gaps in what the field currently focuses on, as the choices of evaluation metrics can profoundly shape the outputs of a field. In most usage contexts, participants cared deeply about uncertainty communication and translucence, but these criteria about exposing the negative aspects of models and explanations are missing in current XAI work. If the field continues focusing solely on model intrinsic properties such as faithfulness, we may risk losing sight of what actually matters for users to achieve their objectives. We also contend that the negligence of diverse usage contexts of XAI can lead to blind spots in the field's focuses. For example, while it is true that faithfulness is paramount for model debugging, which is the use case that XAI researchers tend to fixate on, our results suggest that users are willing to compromise it to a degree for efficient comprehensibility for other usage contexts such as capability assessment. Furthermore, our results reveal nuanced differences in how end-users and experts in the field perceive some of the criteria, specifically pointing to blind spots in what users desire from personalization and interactivity.

\noindent\textbf{Limitations and future directions} The contribution of our study is made by soliciting people's opinions to illustrate patterns of varying relative importance of evaluation criteria. The quantification should be taken with caution. First, our sample size is relatively small. Second, the retrospective approach by soliciting opinions, while allowing exploration of the large problem space, may not reflect the effect size in actual perceptions and behaviors. Future work should explore complementary approaches, such as experimental studies, to further formalize the quantification of their relative importance. We further acknowledge that by synthesizing the list of evaluation criteria from current literature, our work does not explicate their underlying requirements, technical feasibility,  relations between these criteria, or how they are perceived by end-users, which should be explored in future work. Our study also only begins to touch on how to measure these criteria, and we invite future work to explore measurements through simplified or proxy evaluation tasks. 

\section{Acknowledgements}
We wish to thank all participants for their generous time and inputs. We also wish to thank all reviewers for their thoughtful feedback. The majority of the work was completed while the first and second authors were working at IBM Research. FDV also received support from NSF IIS-2107391.

\bibliography{sample-base}
\clearpage
%\clearpage
\appendix

\newpage

%\begin{figure*}
%  \centering
%  \includegraphics[width=0.9\columnwidth]{rank.png}
%   \vspace{-0.4em}
%  \caption{ranking part screenshot (caption TBD)}~\label{fig:rank}

%\end{figure*}
\newpage
\section{Appendix A: Survey Page Screenshot and Descriptions of AI Applications Used }
\label{app-A}

\begin{figure}[!h]
  \centering
  \includegraphics[width=1\columnwidth]{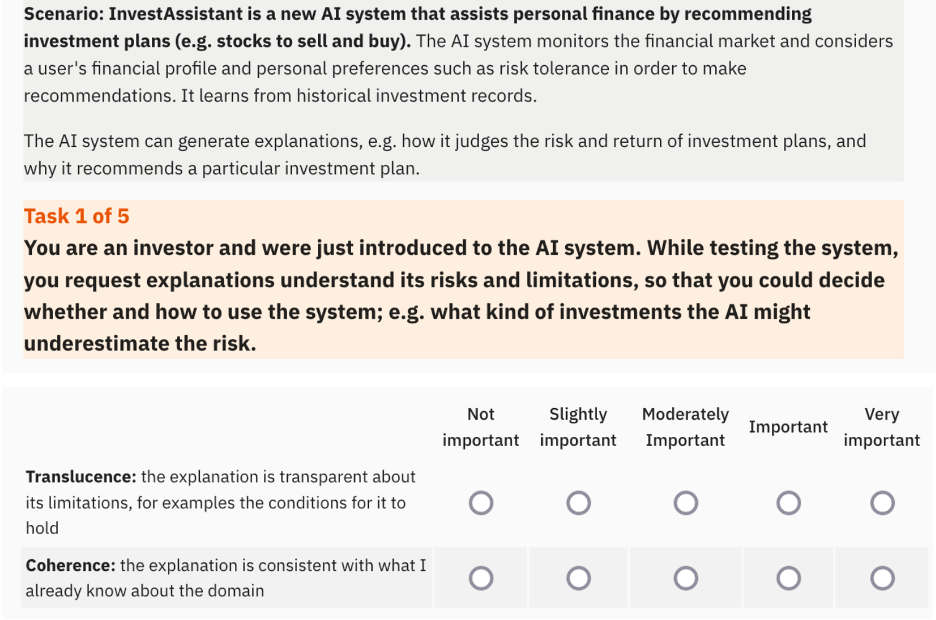}
   \vspace{-0.5em}
  \caption{Screenshot of a survey page, including descriptions of the AI application scenario, one task (for one XAI context), and rating questions for the evaluation criteria (cutting off the other 10 criteria)}~\label{fig:survey}

\end{figure}

\paragraph{AI for personal investment:} InvestAssistant is a new AI system that assists personal finance by recommending investment plans (e.g. stocks to sell and buy). The AI system monitors the financial market and considers a user's financial profile and personal preferences such as risk tolerance in order to make recommendations. It learns from historical investment records. The AI system can generate explanations, e.g. how it judges the risk and return of investment plans, and why it recommends a particular investment plan.

\paragraph{AI for loan application evaluation:} LendingCorp introduces an AI system to help loan officers deciding on whether or not to grant an applicant a loan. The AI system takes information from the loan application form and the applicant's profile, such as the requested loan amount, salary range, and credit score, and predicts the applicant to have high-risk or low-risk of defaulting on the loan. It learns from previous loan applications' information and payment records. The AI system can generate explanations, e.g. how it judges the default risks of applicants, why it predicts a particular applicant to be high-risk or low-risk.

\paragraph{AI for hiring support:} BigCorp introduces an AI system to help hiring managers and teams select candidates to interview for open positions.  The system works by processing resumes of applicants and assessing the fit for an open position based on criteria in the job posting and the hiring team’s preferences. It learns from previous hiring records. The AI system can generate explanations, e.g., how it evaluates the fit of candidates, and why it recommends a particular candidate.

\paragraph{AI for medical diagnosis:} 
HomeDoctor is a new AI system for automated diagnosis. It helps people identify common diseases and suggests the right medical professionals to consult. The system works by examining a user's self-reported symptoms and available physiological measures. It learns from previous diagnostic records and a medical knowledge database. The AI system can generate explanations, e.g. how it makes diagnostic decisions, and why it suggests a particular diagnosis. 

\paragraph{AI for recidivism risk:}An AI system is introduced to help judges make decisions (e.g., sentencing or bail) by assessing a defendant's potential recidivism (re-offending) risk. The system considers the defendant's criminal history and profile information such as employment status, community ties, etc. It learns from the historical data of defendants and their recidivism records. The AI system can generate explanations, e.g., how it assesses recidivism risks, or why it predicts a particular defendant to have high recidivism risk.

\section{Appendix B: Training Tasks}
\label{app-b}
Each screenshot presents one training task to introduce the end-user survey participants to the definition of one evaluation criterion. Participants were given the definition and two explanations, and asked to select which one is better with regard to the given criterion. After submitting their response, the page indicates the correct answer and elaborates on the reason by highlighting the corresponding part of the explanation, as shown in these screenshots.
\\
\\
\\
\\
\begin{figure}[htbp]
  \centering
   \caption{Training page for faithfulness}
  \includegraphics[width=1\columnwidth]{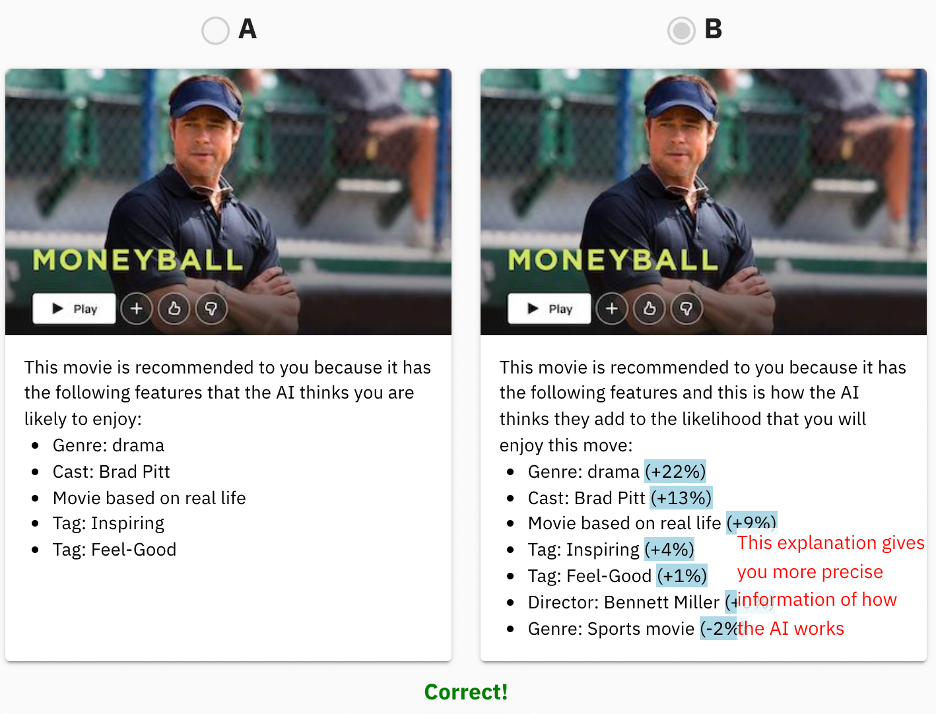}
   \vspace{-0.4em}
\end{figure}

\begin{figure}[htbp]
  \centering
   \caption{Training page for completeness}
  \includegraphics[width=1\columnwidth]{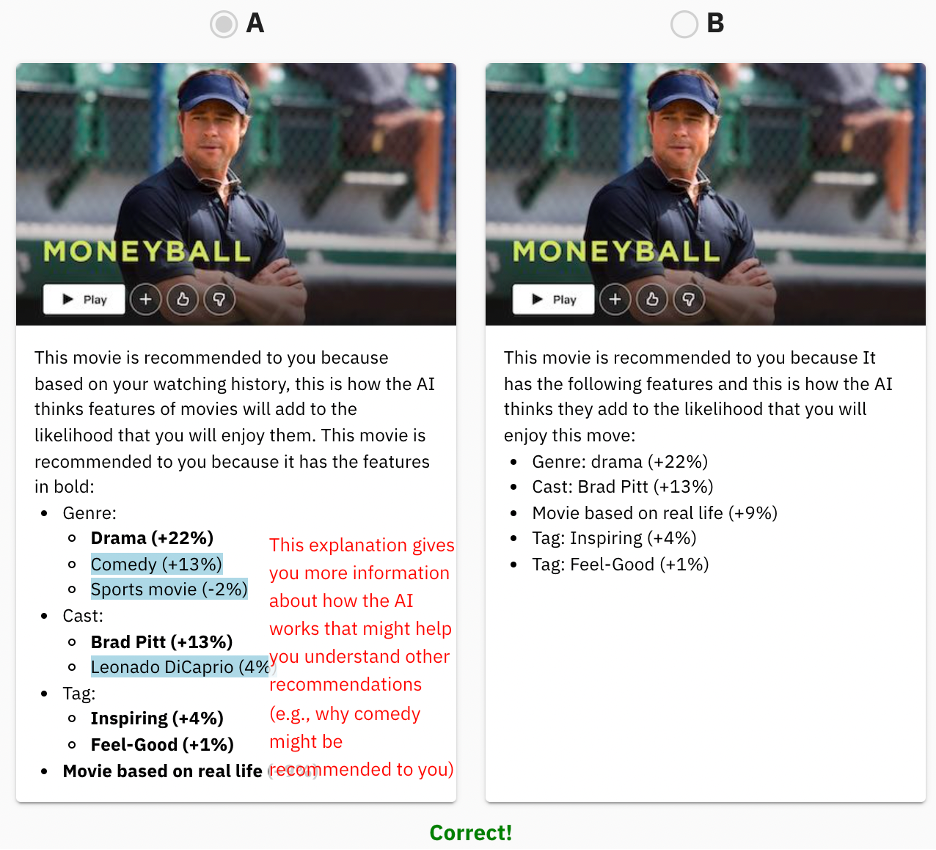}
   \vspace{-0.4em}
\end{figure}

\begin{figure}[htbp]
  \centering
   \caption{Training page for stability}
  \includegraphics[width=1\columnwidth]{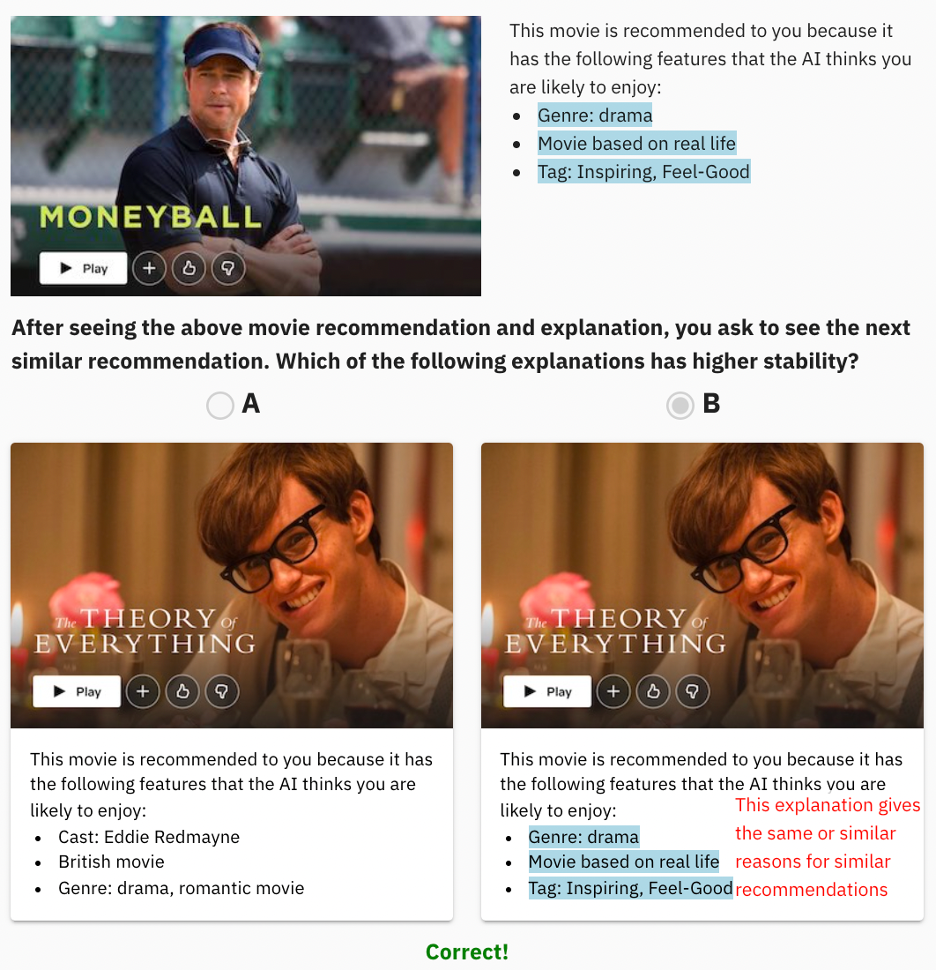}
   \vspace{-0.4em}
\end{figure}

\begin{figure}[htbp]
  \centering
   \caption{Training page for compactness}
  \includegraphics[width=1\columnwidth]{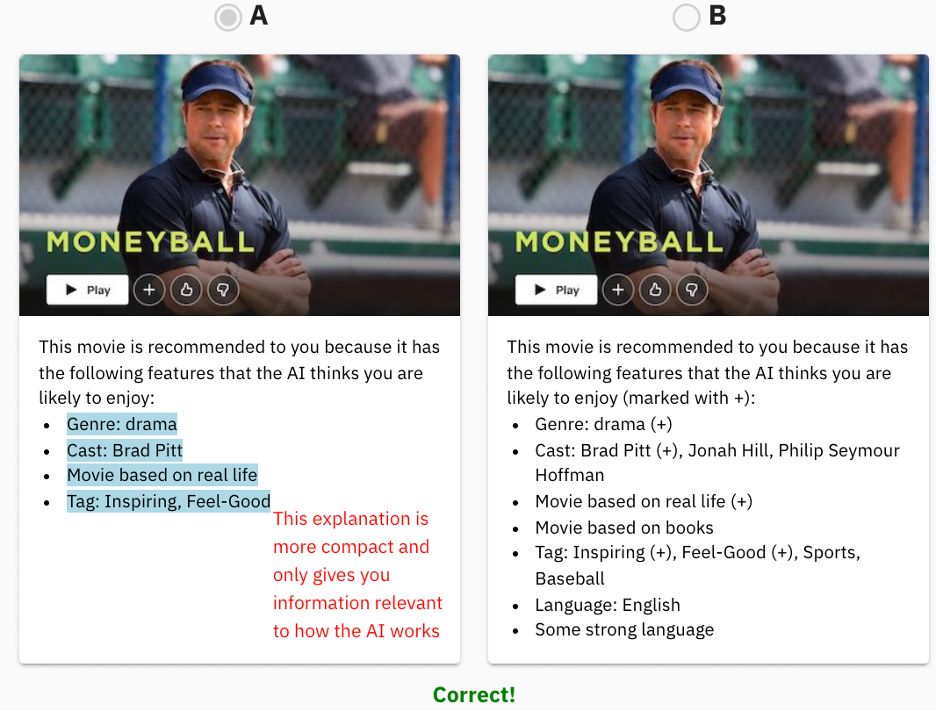}
   \vspace{-0.4em}
\end{figure}

\begin{figure}[htbp]
  \centering
   \caption{Training page for (un)certainty communication}
  \includegraphics[width=1\columnwidth]{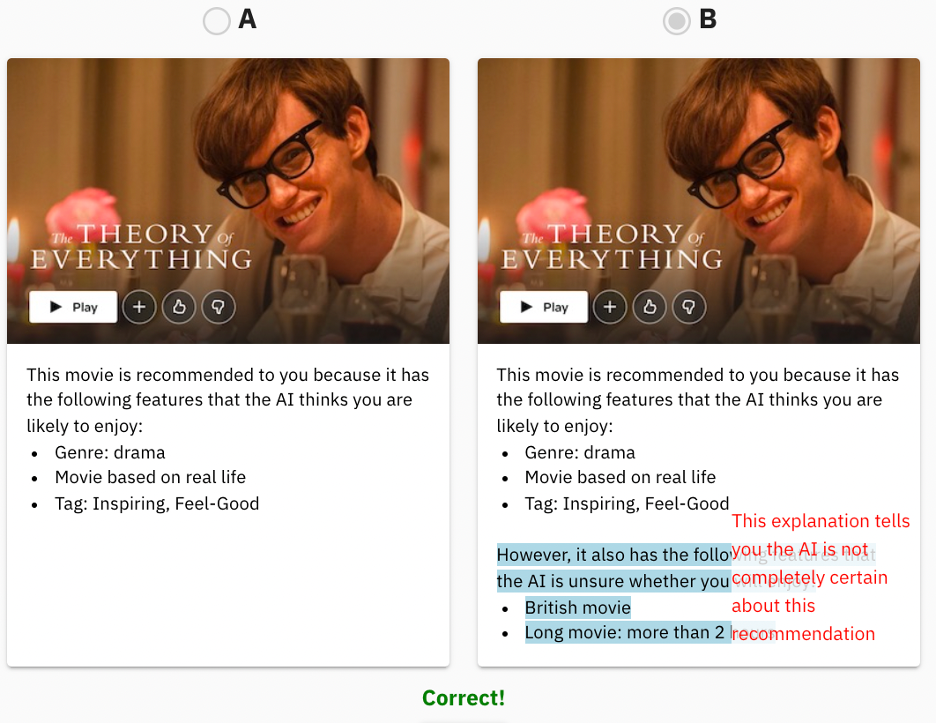}
   \vspace{-0.4em}
\end{figure}

\begin{figure}[htbp]
  \centering
   \caption{Training page for interactivity}
  \includegraphics[width=1\columnwidth]{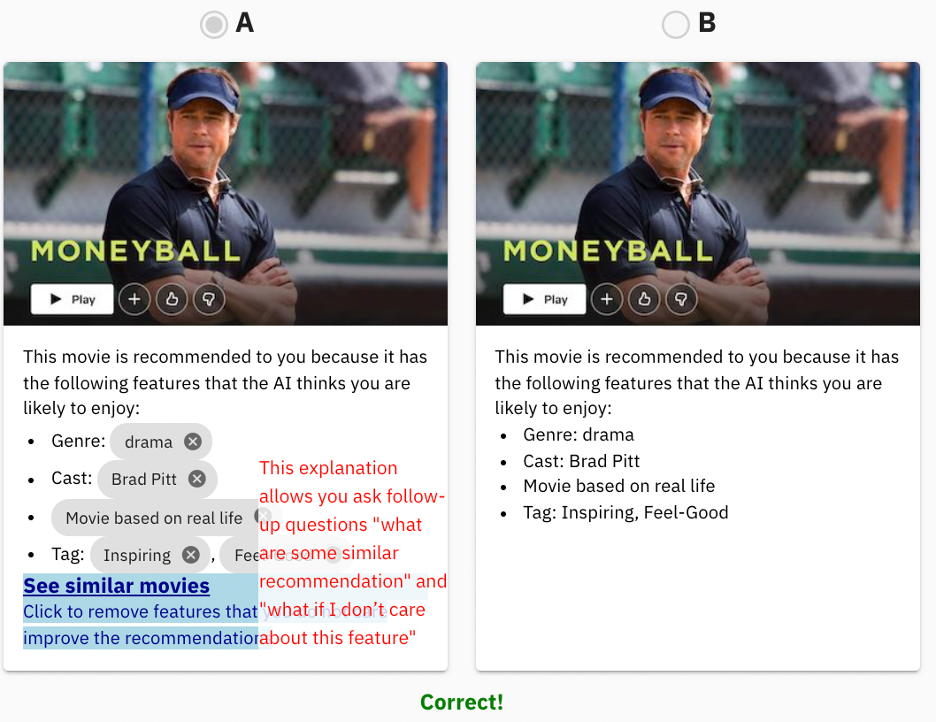}
   \vspace{-0.4em}
\end{figure}

\begin{figure}[htbp]
  \centering
   \caption{Training page for translucence}
  \includegraphics[width=1\columnwidth]{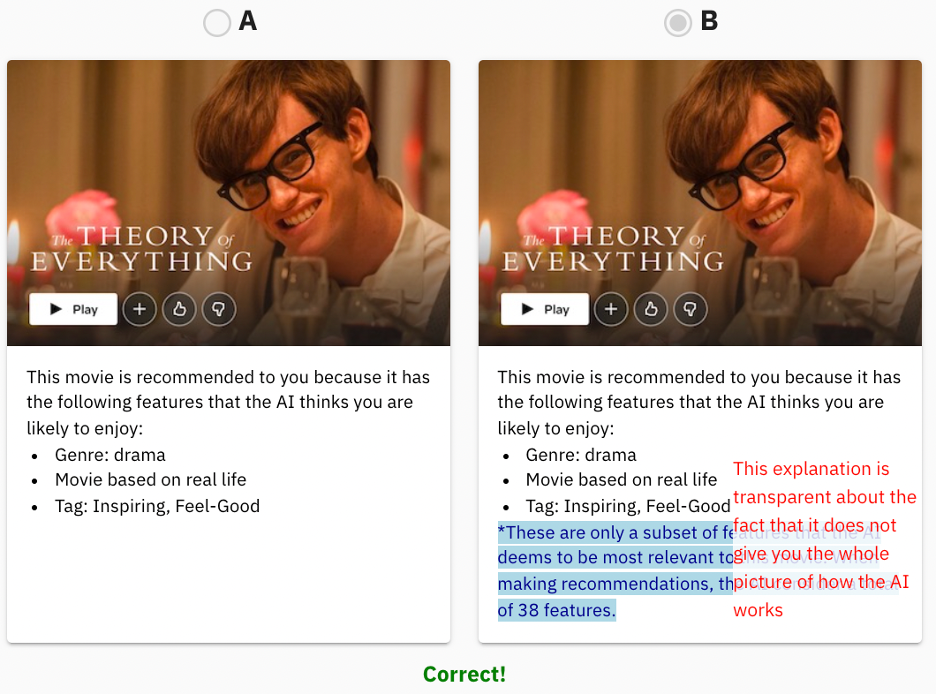}
   \vspace{-0.4em}
\end{figure}

\begin{figure}[htbp]
  \centering
   \caption{Training page for comprehensibility}
  \includegraphics[width=1\columnwidth]{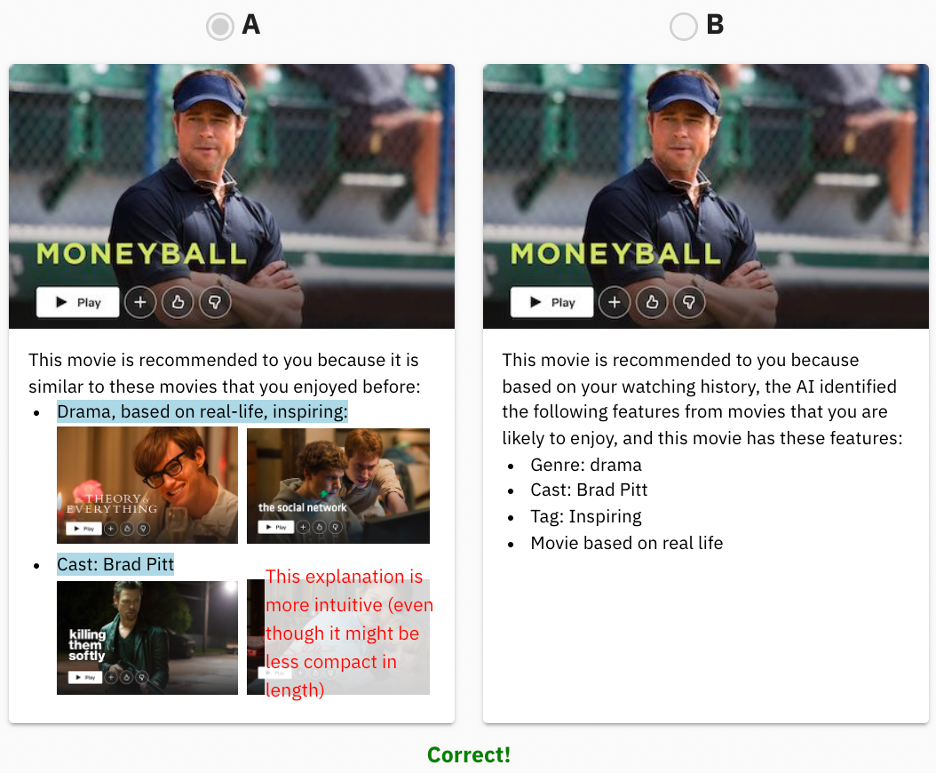}
   \vspace{-0.4em}
\end{figure}

\begin{figure}[htbp]
  \centering
   \caption{Training page for actionability}
  \includegraphics[width=1\columnwidth]{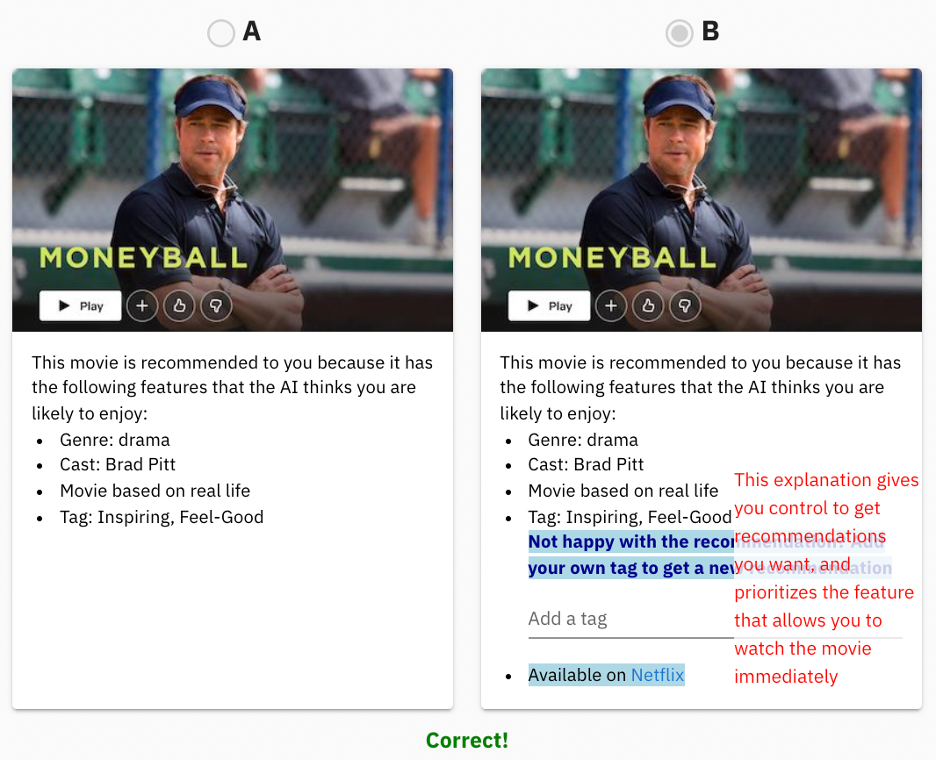}
   \vspace{-0.4em}
\end{figure}

\begin{figure}[htbp]
  \centering
   \caption{Training page for coherence}
  \includegraphics[width=1\columnwidth]{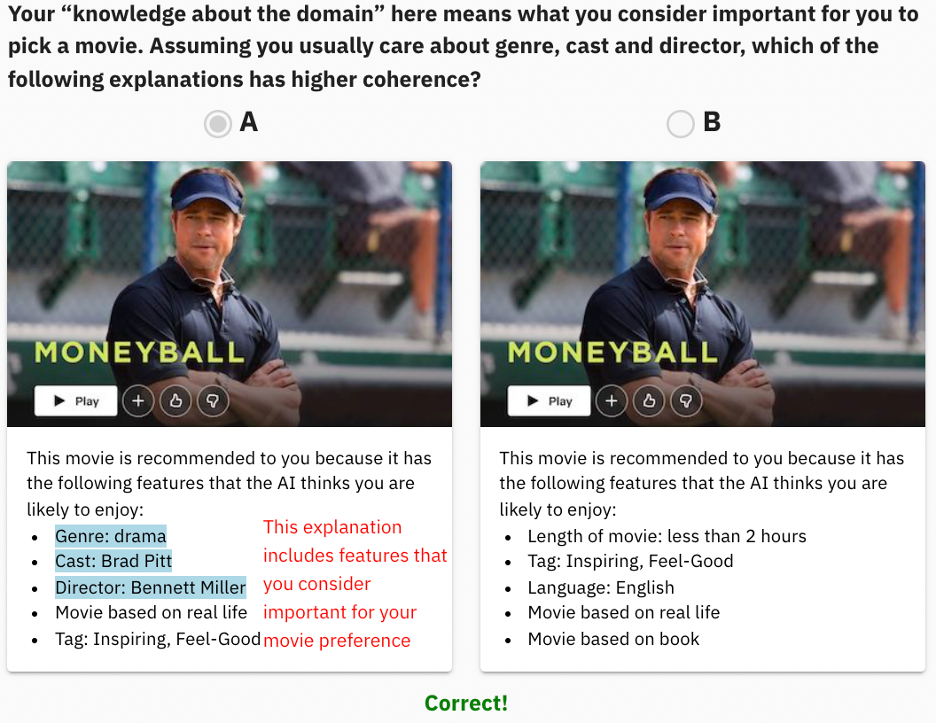}
   \vspace{-0.4em}
\end{figure}

\begin{figure}[htbp]
  \centering
   \caption{Training page for novelty}
  \includegraphics[width=1\columnwidth]{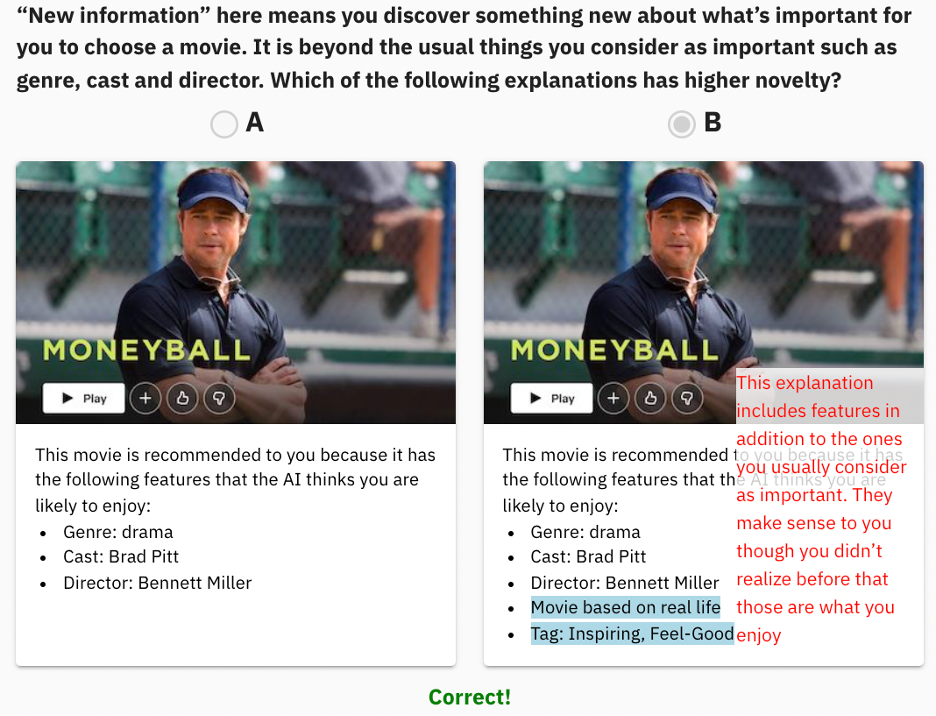}
   \vspace{-0.4em}
\end{figure}

\begin{figure}[htbp]
  \centering
   \caption{Training page for personalization}
  \includegraphics[width=1\columnwidth]{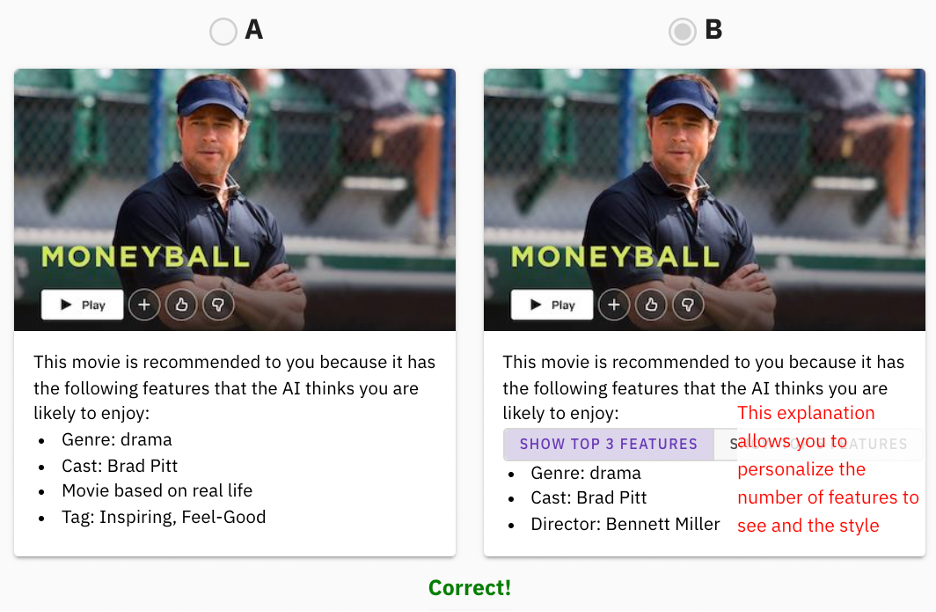}
   \vspace{-0.4em}
\end{figure}

\onecolumn
\section{Appendix C: Participants Information}
\label{app-c}

\begin{table*}[!h]
    \centering
    \caption{Participants information based on the exit survey}
    \begin{tabular}{|c|c|c|}
    \hline
         & XAI experts survey participants & End-user survey participants \\ \hline
         Total N& 35&32\\\hline
         XAI knowledge:&&\\
         \textit{very knowledgeable}&22&0\\
         \textit{knowledgeable}&11&5\\
        \textit{somewhat knowledgeable}&2&16\\
        \textit{no knowledgeable}&0&11\\\hline
        
        Confidence in addressing the AI application:&&\\
        \textit{very confident}&5&4\\
        \textit{confident}&12&15\\
        \textit{somewhat confident}&13&12\\
        \textit{slightly confident}&5&1\\
        \textit{not confident}&0&0\\\hline
        Gender:&&\\
        \textit{female}&10&14\\
        \textit{male}&23&18\\
        \textit{other}&2&0\\\hline
        Age:&&\\
        \textit{under 30}&8&4\\
        \textit{30-39}&15&16\\
        \textit{40-49}&1&7\\
        \textit{50-59}&6&2\\
        \textit{above 60}&2&3\\\hline
        Job titles:& \makecell{32 researchers, \\3 data scientists}&\makecell{8 managers, \\4 software engineers, \\3 data scientists, \\2 designer, \\1 researchers, \\14 as “other”}\\
        \hline

    \end{tabular}

    \label{tab:my_label}
\end{table*}{}

\section{Appendix D: Statistics of post-hoc pairwise comparisons}
\label{app-d}

\begin{table*}[htbp] 
\small
\centering
\caption{Experts Survey:P-Values from pair-wise comparisons of criterion importance ratings for Overall Results (merging 6 contexts)}
\vspace{.5in}
\begin{tabular}{|c|c|c|c|c|c|c|c|c|c|c|c|}
\multicolumn{1}{c|}{}& \begin{rotate}{60} Fidelity \end{rotate} & \begin{rotate}{60} Translucence\end{rotate} & \begin{rotate}{60} Certainty\end{rotate} & \begin{rotate}{60} Stability\end{rotate} & \begin{rotate}{60} Interactivity\end{rotate} & \begin{rotate}{60} Comprehensibility\end{rotate} & \begin{rotate}{60} Completeness\end{rotate} & \begin{rotate}{60} Actionability\end{rotate} & \begin{rotate}{60} Compactness\end{rotate} & \begin{rotate}{60} Coherence\end{rotate} & \begin{rotate}{60} Personalization\end{rotate} \\ \hline
Translucence &0.5801 &- &- &- &- &- &- &- &- &- &- \\ \hline
Certainty &0.0856 &0.1937 &- &- &- &- &- &- &- &- &- \\ \hline
Stability &0.0040 &0.0219 &0.1052 &- &- &- &- &- &- &- &- \\ \hline
Interactivity &0.0002 &0.0004 &0.0093 &0.3795 &- &- &- &- &- &- &- \\ \hline
Comprehensibility &0.0002 &0.0007 &0.0103 &0.2826 &0.8203 &- &- &- &- &- &- \\ \hline
Completeness &<.0001 &<.0001 &<.0001 &0.0060 &0.0565 &0.0769 &- & -& -&- &- \\ \hline
Actionability &<.0001 &<.0001 &<.0001 &0.0006 &0.0020 &0.0023 &0.2771 &- &- &- &- \\ \hline
Compactness &<.0001 &<.0001 &<.0001 &<.0001 &0.0001 &<.0001 &0.0380 &0.3954&- &- &- \\ \hline
Coherence &<.0001 &<.0001 &<.0001 &<.0001 &<.0001 &<.0001 &<.0001 &0.0093 &0.0457 &- &- \\ \hline
Personalization &<.0001 &<.0001 &<.0001 &<.0001 &<.0001 & <.0001 &0.0001& 0.0005& 0.0032& 0.4767&- \\ \hline
Novelty & <.0001& <.0001& <.0001&<.0001 &<.0001 &<.0001 &<.0001 &<.0001 &<.0001 &0.0010 &0.0124 \\ \hline
\end{tabular}
\end{table*}

\begin{table*}[htbp] 
\small
\centering
\vspace{.2in}
\caption{Experts Survey: P-Values from pair-wise comparisons of criterion importance ratings for Model Improvement context}
\vspace{.5in}
\begin{tabular}{|c|c|c|c|c|c|c|c|c|c|c|c|}
\multicolumn{1}{c|}{}& \begin{rotate}{60} Fidelity \end{rotate} & \begin{rotate}{60} Stability\end{rotate} & \begin{rotate}{60} Certainty\end{rotate} & \begin{rotate}{60} Translucense\end{rotate} & \begin{rotate}{60} Comprehensibility\end{rotate} & \begin{rotate}{60} Completeness\end{rotate} & \begin{rotate}{60} Actionability\end{rotate} & \begin{rotate}{60} Compactness\end{rotate} & \begin{rotate}{60} Interactivity\end{rotate} & \begin{rotate}{60} Coherence\end{rotate} & \begin{rotate}{60} Novelty\end{rotate} \\ \hline
Stability &0.1399 &- &- &- &- &- &- &- &- &- &- \\ \hline
Certainty &0.0864 &0.8560 &- &- &- &- &- &- &- &- &- \\ \hline
Translucense &0.0172 &0.4458 &0.4571 &- &- &- &- &- &- &- &- \\ \hline
Comprehensibility &0.0176 &0.2539 &0.2746 &0.6752 &- &- &- &- &- &- &- \\ \hline
Completeness &<.0001 &0.0106 &0.0075 &0.0974 &0.3657 &- &- &- &- &- &- \\ \hline
Actionability &<.0001 &0.0047 &0.0106 &0.0486 &0.1400 &0.5468 &- & -& -&- &- \\ \hline
Compactness &0.0004 &0.0082 &0.0075 &0.0575 &0.0126 &0.4575 &0.8439 &- &- &- &- \\ \hline
Interactivity &<.0001 &0.0001 &<.0001 &0.0049 &0.0284 &0.1790 &0.6030 &0.7465 &- &- &- \\ \hline
Coherence &<.0001 &<.0001 &<.0001 &0.0030 &0.0026 &0.0631 &0.2785 &0.4458 &0.5319 &- &- \\ \hline
Novelty &<.0001 &<.0001 &<.0001 &0.0004 &0.0009 &0.0148 &0.0575 &0.0870 &0.0600 &0.2539 & -\\ \hline
Personalization &<.0001 &<.0001 &<.0001 &0.0002 &<.0001 &0.0047 &0.0172 &0.0036 &0.0255 &0.0864 &0.4710 \\ \hline
\end{tabular}
\end{table*}

\begin{table*}[htbp] 
\small
\centering
\vspace{.2in}
\caption{Experts Survey: P-Values from pair-wise comparisons of criterion importance ratings for Capability Assessment context}
\vspace{.5in}
\begin{tabular}{|c|c|c|c|c|c|c|c|c|c|c|c|}
\multicolumn{1}{c|}{}& \begin{rotate}{60} Translucence \end{rotate} & \begin{rotate}{60} Certainty\end{rotate} & \begin{rotate}{60} Comprehensibility\end{rotate} & \begin{rotate}{60} Fidelity\end{rotate} & \begin{rotate}{60} Stability\end{rotate} & \begin{rotate}{60} Compactness\end{rotate} & \begin{rotate}{60} Interactivity\end{rotate} & \begin{rotate}{60} Actionability\end{rotate} & \begin{rotate}{60} Personalization\end{rotate} & \begin{rotate}{60} Coherence\end{rotate} & \begin{rotate}{60} Completeness\end{rotate} \\ \hline
Certainty &0.5901 &- &- &- &- &- &- &- &- &- &- \\ \hline
Comprehensibility &0.4785 &0.7647 &- &- &- &- &- &- &- &- &- \\ \hline
Fidelity &0.3735 &0.4989 &0.7647 &- &- &- &- &- &- &- &- \\ \hline
Stability &0.1991 &0.2713 &0.5072 &0.7818 &- &- &- &- &- &- &- \\ \hline
Compactness &0.1490 &0.3317 &0.3317 &0.7818 &1.0000 &- &- &- &- &- &- \\ \hline
Interactivity &0.0413 &0.1341 &0.2152 &0.4164 &0.5122& 0.5122 &- & -& -&- &- \\ \hline
Actionability &0.0413 &0.0765 &0.2018 &0.3121 &0.3864 &0.4045 &0.7647 &- &- &- &- \\ \hline
Personalization &0.0130 &0.0657 &0.0917 &0.2969 &0.2713 &0.2677 &0.6440 &0.8220 &- &- &- \\ \hline
Coherence &0.0018 &0.0172 &0.0130 &0.1954 &0.0901 &0.0917 &0.4586& 0.7495& 0.7818 &- &- \\ \hline
Completeness &0.0003 &0.0002 &0.0015 &0.0047 &0.0241 &0.0130 &0.0889 &0.2677 &0.3368 &0.3803 &- \\ \hline
Novelty & 0.5122 & 0.5122 & 0.5122 & 0.5122 & 0.5122 & 0.5122 &0.0002 &0.0006 &0.0004 &0.0013 &0.0241 \\ \hline
\end{tabular}
\end{table*}

\begin{table*}[htbp] 
\small
\centering
\vspace{.2in}
\caption{Experts Survey: P-Values from pair-wise comparisons of criterion importance ratings for Decision Support context}
\vspace{.5in}
\begin{tabular}{|c|c|c|c|c|c|c|c|c|c|c|c|}
\multicolumn{1}{c|}{}& \begin{rotate}{60} Interactivity \end{rotate} & \begin{rotate}{60} Certainty\end{rotate} & \begin{rotate}{60} Fidelity\end{rotate} & \begin{rotate}{60} Translucence\end{rotate} & \begin{rotate}{60} Comprehensibility\end{rotate} & \begin{rotate}{60} Actionability\end{rotate} & \begin{rotate}{60} Stability\end{rotate} & \begin{rotate}{60} Personalization\end{rotate} & \begin{rotate}{60} Compactness\end{rotate} & \begin{rotate}{60} Completeness\end{rotate} & \begin{rotate}{60} Novelty\end{rotate} \\ \hline
Certainty &0.8737 &- &- &- &- &- &- &- &- &- &- \\ \hline
Fidelity &0.5731 &0.6408 &- &- &- &- &- &- &- &- &- \\ \hline
Translucence &0.4555 &0.3858 &0.8737 &- &- &- &- &- &- &- &- \\ \hline
Comprehensibility &0.3375 &0.3136 &0.6408 &0.6438 &- &- &- &- &- &- &- \\ \hline
Acitonability &0.2815 &0.3388 &0.5037 &0.5784 &0.8737 &- &- &- &- &- &- \\ \hline
Stability &0.0902 &0.0601 &0.2815 &0.3189 &0.6732 &0.8541 &- & -& -&- &- \\ \hline
Personalization &0.0187 &0.0219 &0.0515 &0.0484 &0.0986 &0.0933 &0.3249 &- &- &- &- \\ \hline
Compactness &0.0054 &0.0047 &0.0094 &0.0167 &0.0123 &0.0902 &0.1143 &0.7041 &- &- &- \\ \hline
Completeness &0.0041 &0.0041 &0.0041 &0.0047 &0.0047 &0.0319 &0.0674 &0.4288 &0.5635 &- &- \\ \hline
Novelty &0.0016 &0.0047 &0.0049 &0.0094 &0.0484 &0.0368 &0.0931 &0.5335 &0.8134 &0.8645 &- \\ \hline
Coherence &0.0025 &0.0009 &0.0052 &0.0016 &0.0124 &0.0319 &0.0422 &0.4288 &0.7041 &0.8737 &0.9088 \\ \hline
\end{tabular}
\end{table*}

\begin{table*}[htbp] 
\small
\centering
\vspace{.2in}
\caption{Experts Survey: P-Values from pair-wise comparisons of criterion importance ratings for Adapting Control context}
\vspace{.5in}
\begin{tabular}{|c|c|c|c|c|c|c|c|c|c|c|c|}
\multicolumn{1}{c|}{}& \begin{rotate}{60} Certainty \end{rotate} & \begin{rotate}{60} Interactivity\end{rotate} & \begin{rotate}{60} Translucence\end{rotate} & \begin{rotate}{60} Fidelity\end{rotate} & \begin{rotate}{60} Stability\end{rotate} & \begin{rotate}{60} Actionability\end{rotate} & \begin{rotate}{60} Completeness\end{rotate} & \begin{rotate}{60} Comprehensibility\end{rotate} & \begin{rotate}{60} Coherence\end{rotate} & \begin{rotate}{60} Compactness\end{rotate} & \begin{rotate}{60} Personalization\end{rotate} \\ \hline
Interactivity &1.0000 &- &- &- &- &- &- &- &- &- &- \\ \hline
Translucence &0.9141 &0.912 &- &- &- &- &- &- &- &- &- \\ \hline
Fidelity &0.8388 &0.8324 &0.9142 &- &- &- &- &- &- &- &- \\ \hline
Stability &0.1921 &0.2566 &0.2982 &0.3959 &- &- &- &- &- &- &- \\ \hline
Actionability &0.1139 &0.1103 &0.1301 &0.1908 &0.7160 &- &- &- &- &- &- \\ \hline
Completeness &0.0405 &0.0876 &0.0524 &0.0932 &0.4978 &0.9142 &- & -& -&- &- \\ \hline
Comprehensibility &0.0206 &0.0111 &0.0361 &0.0296 &0.2370 &0.4039 &0.5775 &- &- &- &- \\ \hline
Coherence &0.0037 &0.0070 &0.0124 &0.0124 &0.0820& 0.2666&0.2600 &0.6075 &- &- &- \\ \hline
Compactness &0.0021 &0.0014 &0.0021 &0.0021 &0.0462 &0.1196 &0.0836 &0.2104 &0.6889 &- &- \\ \hline
Personalization &0.0004 &<.0001 &0.0001 &0.0005 &0.0092 &0.0091 &0.0251 &0.0293& 0.1908& 0.2702&- \\ \hline
Novelty &0.0001 &<.0001 &0.0002 &<.0001 &0.0023 &0.0234 &0.0092 &0.0386 &0.0890 &0.2425 &0.9462 \\ \hline
\end{tabular}
\end{table*}

\begin{table*}[htbp] 
\small
\centering
\vspace{.2in}
\caption{Experts Survey: P-Values from pair-wise comparisons of criterion importance ratings for Domain Learning context}
\vspace{.5in}
\begin{tabular}{|c|c|c|c|c|c|c|c|c|c|c|c|}
\multicolumn{1}{c|}{}& \begin{rotate}{60} Translucence \end{rotate} & \begin{rotate}{60} Fidelity\end{rotate} & \begin{rotate}{60} Stability\end{rotate} & \begin{rotate}{60} Comprehensibility\end{rotate} & \begin{rotate}{60} Completeness\end{rotate} & \begin{rotate}{60} Interactivity\end{rotate} & \begin{rotate}{60} Certainty\end{rotate} & \begin{rotate}{60} Compactness\end{rotate} & \begin{rotate}{60} Personalization\end{rotate} & \begin{rotate}{60} Actionability\end{rotate} & \begin{rotate}{60} Novelty\end{rotate} \\ \hline
Fidelity &0.9200 &- &- &- &- &- &- &- &- &- &- \\ \hline
Stability &0.8990 &0.9200 &- &- &- &- &- &- &- &- &- \\ \hline
Comprehensibility &0.8177 &0.8990 &0.9200 &- &- &- &- &- &- &- &- \\ \hline
Completeness &0.4102 &0.5338 &0.6247 &0.5904 &- &- &- &- &- &- &- \\ \hline
Interactivity &0.2348 &0.3712 &0.4107 &0.4044 &0.6850 &- &- &- &- &- &- \\ \hline
Certainty &0.0977 &0.1754 &0.2859 &0.3543 &0.5561 &0.9200 &- & -& -&- &- \\ \hline
Compactness &0.0623 &0.0816 &0.1581 &0.0816 &0.2693 &0.4623 &0.5197 &- &- &- &- \\ \hline
Personalization &0.0916 &0.0869 &0.1202 &0.0250 &0.3061 &0.4623 &0.4824 &0.9200 &- &- &- \\ \hline
Actionability &0.0161 &0.0161 &0.0623 &0.0065 &0.0562 &0.1736 &0.0972 &0.4001 &0.4510 &- &- \\ \hline
Novelty &0.0012 &0.0066 &0.0066 &0.0044 &0.0216 &0.0254 &0.0855 &0.2111 &0.3416 &0.8177 &- \\ \hline
Coherence &0.0012 &0.0019 &0.0026 &0.0007 &0.0008 &0.0311 &0.0226 &0.0869 &0.1841 &0.4992 &0.7324 \\ \hline
\end{tabular}
\end{table*}

\begin{table*}[htbp] 
\small
\centering
\vspace{.2in}
\caption{Experts Survey: P-Values from pair-wise comparisons of criterion importance ratings for Model Auditing context}
\vspace{.5in}
\begin{tabular}{|c|c|c|c|c|c|c|c|c|c|c|c|}
\multicolumn{1}{c|}{}& \begin{rotate}{60} Fidelity \end{rotate} & \begin{rotate}{60} Translucence\end{rotate} & \begin{rotate}{60} Completeness\end{rotate} & \begin{rotate}{60} Certainty\end{rotate} & \begin{rotate}{60} Interactivity\end{rotate} & \begin{rotate}{60} Stability\end{rotate} & \begin{rotate}{60} Comprehensibility\end{rotate} & \begin{rotate}{60} Actionability\end{rotate} & \begin{rotate}{60} Compactness\end{rotate} & \begin{rotate}{60} Coherence\end{rotate} & \begin{rotate}{60} Personalization\end{rotate} \\ \hline
Translucence &0.5993 &- &- &- &- &- &- &- &- &- &- \\ \hline
Completeness &0.3756 &0.8191 &- &- &- &- &- &- &- &- &- \\ \hline
Certainty &0.0180 &0.0760 &0.1272 &- &- &- &- &- &- &- &- \\ \hline
Interactivity &0.0133 &0.0399 &0.0733 &0.6810 &- &- &- &- &- &- &- \\ \hline
Stability &0.0104 &0.0507 &0.0570& 0.4410 &0.8512 & - &- &- &- &- &- \\ \hline
Comprehensibility & <.0001&0.0012 &0.0048 &0.0507 &0.2079 &0.2718 &- & -& -&- &- \\ \hline
Actionability &<.0001 &0.0003 &0.0003 &0.0047 &0.0092 &0.0673 &0.1219 &- &- &- &- \\ \hline
Compactness &<.0001 &<.0001 &0.0001 &0.0011 &0.0077 &0.0180 &0.0012 &0.9003 &- &- &- \\ \hline
Coherence &<.0001 &<.0001 &<.0001 &0.0004 &0.0092 &0.0028 &0.0733 &0.8191 &0.8504 &- &- \\ \hline
Personalization &<.0001 &<.0001 &<.0001 &0.0001 &0.0001 &0.0012 &0.0001 &0.1004 &0.0292 &0.1740 &- \\ \hline
Novelty &<.0001 &<.0001 &<.0001 &<.0001 &<.0001 &<.0001 &0.0001 &0.0123 &0.0111 &0.0178 &0.4112 \\ \hline
\end{tabular}
\end{table*}

\begin{table*}[htbp] 
\small
\centering
\vspace{.2in}
\caption{End-User Survey: P-Values from pair-wise comparisons of criterion importance ratings for Overall Results (merging 5 contexts)}
\vspace{.5in}
\begin{tabular}{|c|c|c|c|c|c|c|c|c|c|c|c|}
\multicolumn{1}{c|}{}& \begin{rotate}{60} Fidelity \end{rotate} & \begin{rotate}{60} Certainty\end{rotate} & \begin{rotate}{60} Translucence\end{rotate} & \begin{rotate}{60} Stability\end{rotate} & \begin{rotate}{60} Comprehensibility\end{rotate} & \begin{rotate}{60} Interactivity\end{rotate} & \begin{rotate}{60} Actionability\end{rotate} & \begin{rotate}{60} Personalization\end{rotate} & \begin{rotate}{60} Completeness\end{rotate} & \begin{rotate}{60} Coherence\end{rotate}& \begin{rotate}{60} Novelty\end{rotate} \\ \hline
Certainty &0.2086 &- &- &- &- &- &- &- &- &- &- \\ \hline
Translucence &0.1421 &0.8826 &- &- &- &- &- &- &- &- &- \\ \hline
Stability &0.0114 &0.2563 &0.3606 &- &- &- &- &- &- &- &- \\ \hline
Comprehensibility &0.0013 &0.0540 &0.0601 &0.3463 &- &- &- &- &- &- &- \\ \hline
Interactivity &<.0001 &0.0027 &0.0025 &0.0320 &0.2086 &- &- &- &- &- &- \\ \hline
Actionability &<.0001 &0.0012 &0.0029 &0.0197 &0.1161 &0.8702 &- & -& -&- &- \\ \hline
Personalization &<.0001 &0.0018 &0.0034 &0.0151 &0.1159 &0.7745 &0.8951 &- &- &- &- \\ \hline
Completeness &<.0001 &0.0007 &0.0008 &0.0021 &0.0666 &0.7195& 0.8377& 0.8826& - &- &- \\ \hline
Coherence &<.0001 &<.0001 &<.0001 &<.0001 &0.00047 &0.0944 &0.1150 &0.1235 &0.161 &- &- \\ \hline
Novelty &<.0001 &<.0001 &<.0001 &<.0001 &<.0001 &<.0001 &<.0001 &<.0001 &<.0001 &<.0001 &- \\ \hline
Compactness &<.0001 &<.0001 &<.0001 &<.0001 &<.0001 &<.0001 &<.0001 &<.0001 &<.0001 &<.0001 &0.5993 \\ \hline
\end{tabular}
\end{table*}

\begin{table*}[htbp] 
\small
\centering
\vspace{.2in}
\caption{End-User Survey:P-Values from pair-wise comparisons of criterion importance ratings for Capability Assessment context}
\vspace{.5in}
\begin{tabular}{|c|c|c|c|c|c|c|c|c|c|c|c|}
\multicolumn{1}{c|}{}& \begin{rotate}{60} Comprehensibility \end{rotate} & \begin{rotate}{60} Personalization\end{rotate} & \begin{rotate}{60} Certainty\end{rotate} & \begin{rotate}{60} Fidelity\end{rotate} & \begin{rotate}{60} Stability\end{rotate} & \begin{rotate}{60} Translucence\end{rotate} & \begin{rotate}{60} Actionability\end{rotate} & \begin{rotate}{60} Coherence\end{rotate} & \begin{rotate}{60} Interactivity\end{rotate} & \begin{rotate}{60} Completeness\end{rotate} & \begin{rotate}{60} Compactness\end{rotate} \\ \hline
Personalization &1.0000 &- &- &- &- &- &- &- &- &- &- \\ \hline
Certainty &0.7995 &0.7743 &- &- &- &- &- &- &- &- &- \\ \hline
Fidelity &0.5421 &0.5910 &0.8263 &- &- &- &- &- &- &- &- \\ \hline
Stability &0.5882 &0.5910 &0.8619 &1.0000 &- &- &- &- &- &- &- \\ \hline
Translucence &0.4932 &0.5854 &0.5856 &0.8250 &0.8619 &- &- &- &- &- &- \\ \hline
Actionability &0.1203 &0.0381 &0.4252 &0.4411 &0.4832 &0.6246 &- & -& -&- &- \\ \hline
Coherence &0.0867 &0.1044 &0.4146 &0.3280 &0.2455 &0.5856 &0.8960 &- &- &- &- \\ \hline
Interactivity &0.0281 &0.0035 &0.1544 &0.1097 &0.1875 &0.4253 &0.5566 &0.7623 &- &- &- \\ \hline
Completeness &0.0164 &0.0285 &0.1875 &0.0747 &0.0824 &0.3161 &0.5856 &0.5856 &0.8367 &- &- \\ \hline
Compactness &0.0003 &0.0007 &0.0018 &0.0002 &0.0002 &0.0035 &0.0024 &0.0035 &0.0436 &0.0834 &- \\ \hline
Novelty &<.0001 &<.0001 &0.0018 &0.0002 &0.0002 &0.0035 &0.0024 &0.0035 &0.0043 &0.0132 &0.7131 \\ \hline
\end{tabular}
\end{table*}

\begin{table*}[htbp] 
\small
\centering
\vspace{.2in}
\caption{End-User Survey: P-Values from pair-wise comparisons of criterion importance ratings for Decision Support context}
\vspace{.5in}
\begin{tabular}{|c|c|c|c|c|c|c|c|c|c|c|c|}
\multicolumn{1}{c|}{}& \begin{rotate}{60} Certainty \end{rotate} & \begin{rotate}{60} Translucence\end{rotate} & \begin{rotate}{60} Fidelity\end{rotate} & \begin{rotate}{60} Comprehensibility\end{rotate} & \begin{rotate}{60} Actionability\end{rotate} & \begin{rotate}{60} Stability\end{rotate} & \begin{rotate}{60} Interactivity\end{rotate} & \begin{rotate}{60} Personalization\end{rotate} & \begin{rotate}{60} Coherence\end{rotate} & \begin{rotate}{60} Compactness\end{rotate} & \begin{rotate}{60} Novelty\end{rotate} \\ \hline
Translucence &1.0000 &- &- &- &- &- &- &- &- &- &- \\ \hline
Fidelity &0.7340 &0.7340 &- &- &- &- &- &- &- &- &- \\ \hline
Comprehensibility &0.4919 &0.4919 &0.6414 &- &- &- &- &- &- &- &- \\ \hline
Actionability &0.2382 &0.2541 &0.3476 &0.5403 &- &- &- &- &- &- &- \\ \hline
Stability &0.0630 &0.0566 &0.0798 &0.3076 &0.7340 &- &- &- &- &- &- \\ \hline
Interactivity &0.0515 &0.0515 &0.0969 &0.1813 &0.2891 &0.6989 &- & -& -&- &- \\ \hline
Personalization &0.0300 &0.0187 &0.0548 &0.0778 &0.2223 &0.4919 &0.7340 &- &- &- &- \\ \hline
Coherence &0.0093 &0.0041 &0.0041 &0.0515 &0.2891 &3269 &0.7340 &0.9004 &- &- &- \\ \hline
Compactness &<.0001 &<.0001 &<.0001 &<.0001 &0.0016 &<.0001 &0.0093 &0.0030 &0.0009 &- &- \\ \hline
Novelty &<.0001 &<.0001 &<.0001 &0.0017 &0.0093 &0.0117 &0.0475 &0.0475 &0.0300 &0.3078 &- \\ \hline
Completeness &0.0143 &0.0093 &0.0261 &0.1685 &0.2223 &0.5403 &0.6295 &0.5403 &0.0066 &0.1472 & \\ \hline
\end{tabular}
\end{table*}

\begin{table*}[htbp] 
\small
\centering
\vspace{.2in}
\caption{End-User Survey: P-Values from pair-wise comparisons of criterion importance ratings for Adapting Control context}
\vspace{.5in}
\begin{tabular}{|c|c|c|c|c|c|c|c|c|c|c|c|}
\multicolumn{1}{c|}{}& \begin{rotate}{60} Translucence \end{rotate} & \begin{rotate}{60} Fidelity\end{rotate} & \begin{rotate}{60} Certainty\end{rotate} & \begin{rotate}{60} Stability\end{rotate} & \begin{rotate}{60} Personalization\end{rotate} & \begin{rotate}{60} Actionability\end{rotate} & \begin{rotate}{60} Interactivity\end{rotate} & \begin{rotate}{60} Comprehensibility\end{rotate} & \begin{rotate}{60} Completeness\end{rotate} & \begin{rotate}{60} Coherence\end{rotate} & \begin{rotate}{60} Novelty\end{rotate} \\ \hline
Fidelity &0.7337 &- &- &- &- &- &- &- &- &- &- \\ \hline
Certainty &0.7577 &1.0000 &- &- &- &- &- &- &- &- &- \\ \hline
Stability &0.3619 &0.3621 &0.3619 &- &- &- &- &- &- &- &- \\ \hline
Personalization &0.3285 &0.3619 &0.3619 &0.7577 &- &- &- &- &- &- &- \\ \hline
Actionability &0.2081 &0.2299 &0.2679 &0.5980 &0.7852 &- &- &- &- &- &- \\ \hline
Interactivity &0.0576 &0.1131 &0.1539 &0.4994 &0.5169 &0.7852 &- & -& -&- &- \\ \hline
Comprehensibility &0.0159 &0.0120 &0.0618 &0.2299 &0.4207 &0.4207 &0.5980 &- &- &- &- \\ \hline
Completeness &0.0055 &0.0055 &0.0116 &0.0847 &0.3226 &0.3721 &0.4577 &0.7427 &- &- &- \\ \hline
Coherence &0.0056 &0.0039 &0.0218 &0.0386 &0.1785 &0.2299 &0.3619 &0.5185 &0.7852 &- &- \\ \hline
Novelty &<.0001 &<.0001 &<.0001 &<.0001 &<.0001 &<.0001 &0.0003 &0.0021& 0.0024& 0.0017 &- \\ \hline
Compactness &<.0001 &<.0001 &<.0001 &<.0001 &0.0002 &0.0002& 0.0009& 0.0002& 0.0007& <.0001& 0.7427 \\ \hline
\end{tabular}
\end{table*}

\begin{table*}[htbp] 
\small
\centering
\vspace{.2in}
\caption{End-User Survey: P-Values from pair-wise comparisons of criterion importance ratings for Domain Learning context}
\vspace{.5in}
\begin{tabular}{|c|c|c|c|c|c|c|c|c|c|c|c|}
\multicolumn{1}{c|}{}& \begin{rotate}{60} Stability \end{rotate} & \begin{rotate}{60} Fidelity\end{rotate} & \begin{rotate}{60} Certainty\end{rotate} & \begin{rotate}{60} Comprehensibility\end{rotate} & \begin{rotate}{60} Completeness\end{rotate} & \begin{rotate}{60} Translucence\end{rotate} & \begin{rotate}{60} Personalization\end{rotate} & \begin{rotate}{60} Interactivity\end{rotate} & \begin{rotate}{60} Actionability\end{rotate} & \begin{rotate}{60} Coherence\end{rotate} & \begin{rotate}{60} Compactness\end{rotate} \\ \hline
Fidelity &0.6902 &- &- &- &- &- &- &- &- &- &- \\ \hline
Certainty &0.5498 &0.9207 &- &- &- &- &- &- &- &- &- \\ \hline
Comprehensibility &0.6017 &0.9212 &1.0000 &- &- &- &- &- &- &- &- \\ \hline
Completeness &0.2024 &0.4035 &0.5555 &0.5498 &- &- &- &- &- &- &- \\ \hline
Translucence &0.2746 &0.4674 &0.5090 &0.5386 &1.0000 &- &- &- &- &- &- \\ \hline
Personalization &0.3028 &0.5386 &0.5498 &0.5090 &0.9212 &0.9212 &- & -& -&- &- \\ \hline
Interactivity &0.1804 &0.5090 &0.5386 &0.4947 &0.8841 &0.8841 &0.9207 &- &- &- &- \\ \hline
Actionability &0.0589 &0.1104 &0.1804 &0.1107 &0.4674 &0.4947 &0.4216 &0.5386 &- &- &- \\ \hline
Coherence &0.0186 &0.0322 &0.0492 &0.719 &0.1702 &0.1431 &0.2099 &0.3101 &0.6958 &- &- \\ \hline
Compactness &0.0054 &0.0155 &0.0449 &0.0089 &0.0582 &0.0877 &0.1186 &0.1107 &0.2658 &0.5386 &- \\ \hline
Novelty &0.0006 &0.0006 &0.0006 & 0.0008 &0.0044 &0.0013 &0.0067 &0.0116 &0.0322 &0.0116 &0.2024 \\ \hline
\end{tabular}
\end{table*}

\begin{table*}[htbp] 
\small
\centering
\vspace{.2in}
\caption{End-User Survey: P-Values from pair-wise comparisons of criterion context}
\vspace{.5in}
\begin{tabular}{|c|c|c|c|c|c|c|c|c|c|c|c|}
\multicolumn{1}{c|}{}& \begin{rotate}{60} Fidelity \end{rotate} & \begin{rotate}{60} Translucence\end{rotate} & \begin{rotate}{60} Completeness\end{rotate} & \begin{rotate}{60} Certainty\end{rotate} & \begin{rotate}{60} Stability\end{rotate} & \begin{rotate}{60} Interactivity\end{rotate} & \begin{rotate}{60} Comprehensibility\end{rotate} & \begin{rotate}{60} Actionability\end{rotate} & \begin{rotate}{60} Coherence\end{rotate} & \begin{rotate}{60} Personalization\end{rotate} & \begin{rotate}{60} Novelty\end{rotate} \\ \hline
Translucence &0.0496 &- &- &- &- &- &- &- &- &- &- \\ \hline
Completeness &0.0007 &0.7803 &- &- &- &- &- &- &- &- &- \\ \hline
Certainty &0.0194 &0.4711 &0.8224 &- &- &- &- &- &- &- &- \\ \hline
Stability &0.0031 &0.5805 &0.7803 &1.0000 &- &- &- &- &- &- &- \\ \hline
Interactivity &0.0031 &0.2654 &0.4613 &0.6455 &0.6186 &- &- &- &- &- &- \\ \hline
Comprehensibility &<.0001 &0.0394 &0.0530 &0.0961 &0.0810 &0.2189 &0.8224- & -& -&- &- \\ \hline
Actionability &0.0001 &0.0407 &0.0757 &0.0961 &0.0810 &0.2189 &0.8224 &- &- &- &- \\ \hline
Coherence &<.0001 &0.0114 &0.0059 &0.0129 &0.0221 &0.2189 &0.6079 &0.8224 &- &- &- \\ \hline
Personalization & <.0001& 0.0034& <.0001& 0.0125& 0.0017& 0.0079& 0.0496& 0.1351& 0.1033&- &- \\ \hline
Novelty &<.0001 &0.0003 &<.0001 &0.0028 &0.0005 &0.0019 &0.0961 &0.1033& 0.1202& 0.8224 &- \\ \hline
Compactness &<.0001 &<.0001 &<.0001 &<.0001 &<.0001 &<.0001 &<.0001 &0.0004 &0.0002 &0.0548 &0.0982 \\ \hline
\end{tabular}
\end{table*}

\begin{table*}[htbp] 
\small
\centering
\vspace{.2in}
\caption{Merged Experts and End-User Survey: P-Values from pair-wise comparisons of criterion importance ratings for Overall Results}
\vspace{.5in}
\begin{tabular}{|c|c|c|c|c|c|c|c|c|c|c|c|}
\multicolumn{1}{c|}{}& \begin{rotate}{60} Fidelity \end{rotate} & \begin{rotate}{60} Translucence\end{rotate} & \begin{rotate}{60} Certainty\end{rotate} & \begin{rotate}{60} Stability\end{rotate} & \begin{rotate}{60} Interactivity\end{rotate} & \begin{rotate}{60} Comprehensibility\end{rotate} & \begin{rotate}{60} Completeness\end{rotate} & \begin{rotate}{60} Actionability\end{rotate} & \begin{rotate}{60} Personalization\end{rotate} & \begin{rotate}{60} Coherence\end{rotate} & \begin{rotate}{60} Compactness\end{rotate} \\ \hline
Translucence & 0.5676&- &- &- &- &- &- &- &- &- &- \\ \hline
Certainty &0.0851 &0.1810 &- &- &- &- &- &- &- &- &- \\ \hline
Stability &0.0004 &0.0046 &0.0321 &- &- &- &- &- &- &- &- \\ \hline
Interactivity &<.0001 &<.0001 &0.0024 &0.3042 &- &- &- &- &- &- &- \\ \hline
Comprehensibility &<.0001 &<.0001 &0.0021 &0.2627 &0.9396 &- &- &- &- &- &- \\ \hline
Completeness &<.0001 &<.0001 &<.0001 &0.0006 &0.0248 &0.0172 &- & -& -&- &- \\ \hline
Actionability &<.0001 &<.0001 &<.0001 &0.0004 &0.0018 &0.0018 &0.5423 &- &- &- &- \\ \hline
Personalization &<.0001 &<.0001 &<.0001 &<.0001 &<.0001 &<.0001 &0.0148 &0.0247 &- &- &- \\ \hline
Coherence &<.0001 &<.0001 &<.0001 &<.0001 &<.0001 &<.0001 &<.0001 &0.0036 &0.3694 &- &- \\ \hline
Compactness &<.0001 &<.0001 &<.0001 &<.0001 &<.0001 &<.0001 &<.0001 &<.0001 &0.0004 &0.0038 &- \\ \hline
Novelty &<.0001 &<.0001 &<.0001 &<.0001 &<.0001 &<.0001 &<.0001 &<.0001 &<.0001 &<.0001 &0.0012 \\ \hline
\end{tabular}
\end{table*}

\begin{table*}[htbp] 
\small
\centering
\vspace{.2in}
\caption{Merged Experts and End-User Survey: P-Values from pair-wise comparisons of criterion importance ratings for Capability Assessment context}
\vspace{.5in}
\begin{tabular}{|c|c|c|c|c|c|c|c|c|c|c|c|}
\multicolumn{1}{c|}{}& \begin{rotate}{60} Comprehensibility \end{rotate} & \begin{rotate}{60} Certainty\end{rotate} & \begin{rotate}{60} Tranlucence\end{rotate} & \begin{rotate}{60} Fidelity\end{rotate} & \begin{rotate}{60} Stability\end{rotate} & \begin{rotate}{60} Personalization\end{rotate} & \begin{rotate}{60} Actionability\end{rotate} & \begin{rotate}{60} Interactivity\end{rotate} & \begin{rotate}{60} Coherence\end{rotate} & \begin{rotate}{60} Compactness\end{rotate} & \begin{rotate}{60} Completeness\end{rotate} \\ \hline
Certainty &1.0000 &- &- &- &- &- &- &- &- &- &- \\ \hline
Translucence &1.0000 &1.0000 &- &- &- &- &- &- &- &- &- \\ \hline
Fidelity &0.5136 &0.4722 &0.5103 &- &- &- &- &- &- &- &- \\ \hline
Stability &0.3682 &0.3821 &0.4325 &0.8400 &- &- &- &- &- &- &- \\ \hline
Personalization &0.1499 &0.1731 &0.1836 &0.5103 &0.5579 &- &- &- &- &- &- \\ \hline
Actionability &0.0339 &0.339 &0.0474 &0.1525 &0.2040 &0.3682 &- & -& -&- &- \\ \hline
Interactivity &0.0131 &0.0227 &0.0262 &0.1065 &0.1556 &0.4356 &1.0000 &- &- &- &- \\ \hline
Coherence &0.0012 &0.0132 &0.0079 &0.0654 &0.0227 &0.2306 &0.6889 &0.6627 &- &- &- \\ \hline
Compactness &0.0002 &0.0041 &0.0021 &0.0208 &0.0196 &0.1405 &0.4484 &0.3821 &0.5795 &- &- \\ \hline
Completeness &<.0001 &0.0002 &0.0003 &0.0003 &0.0021 &0.0334 &0.1674 &0.1142 &0.2521 &0.5579 &- \\ \hline
Novelty &<.0001 &<.0001 &<.0001 &<.0001 &<.0001 &<.0001 &<.0001 &<.0001 &<.0001 &0.0002 &0.0003 \\ \hline
\end{tabular}
\end{table*}

\begin{table*}[htbp] 
\small
\centering
\vspace{.2in}
\caption{Merged Experts and End-User Survey: P-Values from pair-wise comparisons of criterion importance ratings for  Decision Support context}
\vspace{.5in}
\begin{tabular}{|c|c|c|c|c|c|c|c|c|c|c|c|}
\multicolumn{1}{c|}{}& \begin{rotate}{60} Certainty \end{rotate} & \begin{rotate}{60} Translucence\end{rotate} & \begin{rotate}{60} Fidelity\end{rotate} & \begin{rotate}{60} Comprehensibility\end{rotate} & \begin{rotate}{60} Interactivity\end{rotate} & \begin{rotate}{60} Actionability\end{rotate} & \begin{rotate}{60} Stability\end{rotate} & \begin{rotate}{60} Personalization\end{rotate} & \begin{rotate}{60} Coherence\end{rotate} & \begin{rotate}{60} Compactness\end{rotate} & \begin{rotate}{60} Novelty\end{rotate} \\ \hline
Translucence &0.5003 &- &- &- &- &- &- &- &- &- &- \\ \hline
Fidelity &0.5117 &0.9077 &- &- &- &- &- &- &- &- &- \\ \hline
Cmoprehensibility &0.1802 &0.3666 &0.4531 &- &- &- &- &- &- &- &- \\ \hline
Interactivity &0.1274 &0.3074 &0.3666 &0.9317 &- &- &- &- &- &- &- \\ \hline
Acitonability &0.1031 &0.2011 &0.2073 &0.5924 &0.6254 &- &- &- &- &- &- \\ \hline
Stability &0.0043 &0.0272 &0.0319 &0.3064 &0.3666 &0.6756 &- & -& -&- &- \\ \hline
Personalization &0.0006 &0.0013 &0.0035 &0.0109 &0.0168 &0.0236 &0.1821 &- &- &- &- \\ \hline
Coherence &<.0001 &<.0001 &<.0001 &0.0006 &0.0041 &0.0105 &0.0150 &0.4383&- &- &- \\ \hline
Compactness &<.0001 &<.0001 &<.0001 &<.0001 &<.0001 &0.0002 &<.0001 &0.0105 &0.1120 &- &- \\ \hline
Novelty &<.0001 &<.0001 &<.0001 &0.0001 &<.0001 &0.0003 &0.0019 &0.0491 &0.1943 &0.6892 &- \\ \hline
Completeness &<.0001 &<.0001 &<.0001 &<.0001 &0.0035 &0.0054 &0.0162 &0.3064 &0.6289 &0.2073 &0.5003 \\ \hline
\end{tabular}
\end{table*}

\begin{table*}[htbp] 
\small
\centering
\vspace{.2in}
\caption{Merged Experts and End-User Survey: P-Values from pair-wise comparisons of criterion importance ratings for Adapting Control context}
\vspace{.5in}
\begin{tabular}{|c|c|c|c|c|c|c|c|c|c|c|c|}
\multicolumn{1}{c|}{}& \begin{rotate}{60} Translucence \end{rotate} & \begin{rotate}{60} Certainty\end{rotate} & \begin{rotate}{60} Fidelity\end{rotate} & \begin{rotate}{60} Interactivity\end{rotate} & \begin{rotate}{60} Stability\end{rotate} & \begin{rotate}{60} Actionability\end{rotate} & \begin{rotate}{60} Completeness\end{rotate} & \begin{rotate}{60} Comprehensiblity\end{rotate} & \begin{rotate}{60} Coherence\end{rotate} & \begin{rotate}{60} Personalization\end{rotate} & \begin{rotate}{60} Compactness\end{rotate} \\ \hline
Certainty &0.8920 &- &- &- &- &- &- &- &- &- &- \\ \hline
Fidelity &0.7665 &0.8073 &- &- &- &- &- &- &- &- &- \\ \hline
Interactivity &0.2022 &0.2136 &0.2807 &- &- &- &- &- &- &- &- \\ \hline
Stability &0.1333 &0.0762 &0.1987 &0.7798 &- &- &- &- &- &- &- \\ \hline
Actionability &0.0300 &0.0337 &0.0480 &0.2584 &0.4735 &- &- &- &- &- &- \\ \hline
Completeness &0.0004 &0.0005 &0.0010 &0.0585 &0.0727 &0.4529 &- & -& -&- &- \\ \hline
Comprehensibility &0.0007 &0.0010 &0.0005 &0.0176 &0.0622 &0.2109 &0.8073 &- &- &- &- \\ \hline
Coherence &0.0001 &<.0001 &<.0001 &0.0041 &0.0038 &0.0762 &0.2675 &0.3716 &- &- &- \\ \hline
Personalization &0.0002 &0.0002 &0.0003 &0.0008 &0.0123 &0.0312 &0.2807 &0.3063 &0.8073 &- &- \\ \hline
Compactness &<.0001 &<.0001 &<.0001 &<.0001 &<.0001 &0.0002 &0.0002 &0.0002 &0.0064 &0.0831 &- \\ \hline
Novelty &<.0001 &<.0001 &<.0001 &<.0001 &<.0001 &<.0001 &<.0001 &0.0002 &0.0005 &0.0129 &0.4227 \\ \hline
\end{tabular}
\end{table*}

\begin{table*}[htbp] 
\small
\centering
\vspace{.2in}
\caption{Merged Experts and End-User Survey: P-Values from pair-wise comparisons of criterion importance ratings for Domain Learning context}
\vspace{.5in}
\begin{tabular}{|c|c|c|c|c|c|c|c|c|c|c|c|}
\multicolumn{1}{c|}{}& \begin{rotate}{60} Stability \end{rotate} & \begin{rotate}{60} Fidelity\end{rotate} & \begin{rotate}{60} Comprehensibility\end{rotate} & \begin{rotate}{60} Translucence\end{rotate} & \begin{rotate}{60} Completeness\end{rotate} & \begin{rotate}{60} Certainty\end{rotate} & \begin{rotate}{60} Interactivity\end{rotate} & \begin{rotate}{60} Personalization\end{rotate} & \begin{rotate}{60} Actionability\end{rotate} & \begin{rotate}{60} Compactness\end{rotate} & \begin{rotate}{60} Coherence\end{rotate} \\ \hline
Fidelity &0.8686 &- &- &- &- &- &- &- &- &- &- \\ \hline
Comprehensibility &0.7175 &0.8305 &- &- &- &- &- &- &- &- &- \\ \hline
Translucence &0.6121 &0.6515 &0.8727 &- &- &- &- &- &- &- &- \\ \hline
Completeness &0.2453 &0.2733 &0.3771 &0.5020 &- &- &- &- &- &- &- \\ \hline
Certainty &0.1850 &0.2041 &0.3771 &0.3771 &0.9366 &- &- &- &- &- &- \\ \hline
Interactivity &0.1146 &0.2041 &0.2082 &0.2618 &0.6234 &0.6658 &- & -& -&- &- \\ \hline
Personalization &0.0376 &0.0525 &0.0169 &0.1222 &0.3069 &0.3069 &0.5094 &- &- &- &- \\ \hline
Actionability &0.0041 &0.0010 &0.0004 &0.0076 &0.0224 &0.0169 &0.0979 &0.2178 &- &- &- \\ \hline
Compactness &0.0016 &0.0010 &0.0006 &0.0038 &0.0189 &0.0376 &0.0703 &0.2733 &1.0000 &- &- \\ \hline
Coherence &<.0001 &<.0001 &<.0001 &<.0001 &<.0001 &0.0007 &0.0083 &0.0447 &0.3771 &0.3771 &- \\ \hline
Novelty &<.0001 &<.0001 &<.0001 &<.0001 &<.0001 &<.0001 &0.0002 &0.0048 &0.0977 &0.0428 &0.2844 \\ \hline
\end{tabular}
\end{table*}

\begin{table*}[htbp] 
\small
\centering
\vspace{.2in}
\caption{Merged Experts and End-User Survey: P-Values from pair-wise comparisons of criterion importance ratings for Model Auditing context}
\vspace{.5in}
\begin{tabular}{|c|c|c|c|c|c|c|c|c|c|c|c|}
\multicolumn{1}{c|}{}& \begin{rotate}{60} Fidelity \end{rotate} & \begin{rotate}{60} Translucence\end{rotate} & \begin{rotate}{60} Completeness\end{rotate} & \begin{rotate}{60} Certainty\end{rotate} & \begin{rotate}{60} Stability\end{rotate} & \begin{rotate}{60} Interactivity\end{rotate} & \begin{rotate}{60} Comprehensibility\end{rotate} & \begin{rotate}{60} Actionability\end{rotate} & \begin{rotate}{60} Coherence\end{rotate} & \begin{rotate}{60} Personalization\end{rotate} & \begin{rotate}{60} Compactness\end{rotate} \\ \hline
Translucence &0.0510 &- &- &- &- &- &- &- &- &- &- \\ \hline
Completeness &0.0030 &0.6752 &- &- &- &- &- &- &- &- &- \\ \hline
Certainty &0.0004 &0.0510 &0.1945 &- &- &- &- &- &- &- &- \\ \hline
Stability &<.0001 &0.0457 &0.0618 &0.5796 &- &- &- &- &- &- &- \\ \hline
Interactivity &<.0001 &0.0178 &0.0510 &0.5014 &0.8253 &- &- &- &- &- &- \\ \hline
Comprehensibility &<.0001 &<.0001 &0.0003 &0.0067 &0.0510 &0.1241 &- & -& -&- &- \\ \hline
Actionability &<.0001 &<.0001 &<.0001 &0.0006 &0.0101 &0.0031 &0.1711 &- &- &- &- \\ \hline
Coherence &<.0001 &<.0001 &<.0001 &<.0001 &v &0.0032 &0.0602 &0.7273 &- &- &- \\ \hline
Personalization &<.0001 &<.0001 &<.0001 &<.0001 &<.0001 &<.0001 &<.0001 &0.0211 &0.0313 &- &- \\ \hline
Compactness &<.0001 &<.0001 &<.0001 &<.0001 &<.0001 &<.0001 &<.0001 &0.0067 &0.0230 &0.8567 &- \\ \hline
Novelty &<.0001 &<.0001 &<.0001 &<.0001 &<.0001 &<.0001 &v &0.0024 &0.0040 &0.4712 &0.6098 \\ \hline
\end{tabular}
\end{table*}

\end{document}